\documentclass[lettersize,journal]{IEEEtran}
\usepackage{amsmath,amsfonts}
\usepackage{algorithm}
\usepackage{algorithmicx}
\usepackage{algpseudocode} 
\usepackage{array}
\usepackage{textcomp}
\usepackage{stfloats}
\usepackage{url}
\usepackage{verbatim}
\usepackage{graphicx}
\usepackage{cite}
\usepackage{booktabs} 
\usepackage{multirow}

\usepackage{subfigure}

\usepackage{color}

\begin{document}

\title{DA-SPS: A Dual-stage Network based on Singular Spectrum Analysis, Patching-strategy and Spearman-correlation for Multivariate Time-series Prediction}

\author{Tianhao Zhang*, Shusen Ma*, Yu Kang,~\IEEEmembership{Senior Member,~IEEE,} and Yun-Bo Zhao$^{\dag}$,~\IEEEmembership{Senior Member,~IEEE}
\thanks{This work was supported by the National Natural Science Foundation of China (No. 62173317).}
\thanks{Tianhao Zhang* (equal contribution) is with the Department of Automation, University of Science and Technology of China, Hefei, Anhui, China.}
\thanks{Shusen Ma* (equal contribution) is with the Institute of Advanced Technology, University of Science and Technology of China, Hefei, Anhui, China.}
\thanks{Yu Kang is with the Institute of Advanced Technology, University of Science and Technology of China, Hefei, Anhui, China, the Department of Automation, University of Science and Technology of China, Hefei, Anhui, China, and also with the Institute of Artificial Intelligence, Hefei Comprehensive National Science Center, Anhui, China.}
\thanks{Yun-Bo Zhao$^{\dag}$ (corresponding author) is with the Institute of Advanced Technology, University of Science and Technology of China, Hefei, Anhui, China, the Department of Automation, University of Science and Technology of China, Hefei, Anhui, China, and also with the Institute of Artificial Intelligence, Hefei Comprehensive National Science Center, Anhui, China (e-mail: ybzhao@ustc.edu.cn).}
}



\maketitle

\begin{abstract}
Multivariate time-series forecasting, as a typical problem in the field of time series prediction, has a wide range of applications in weather forecasting, traffic flow prediction, and other scenarios. However, existing works do not effectively consider the impact of extraneous variables on the prediction of the target variable. On the other hand, they fail to fully extract complex sequence information based on various time patterns of the sequences. To address these drawbacks, we propose a DA-SPS model, which adopts different modules for feature extraction based on the information characteristics of different variables. DA-SPS mainly consists of two stages: the target variable processing stage (TVPS) and the extraneous variables processing stage (EVPS). In TVPS, the model first uses Singular Spectrum Analysis (SSA) to process the target variable sequence and then uses Long Short-Term Memory (LSTM) and P-Conv-LSTM which deploys a patching strategy to extract features from trend and seasonality components, respectively. In EVPS, the model filters extraneous variables that have a strong correlation with the target variate by using Spearman correlation analysis and further analyses them using the L-Attention module which consists of LSTM and attention mechanism. Finally, the results obtained by TVPS and EVPS are combined through weighted summation and linear mapping to produce the final prediction. The results on four public datasets demonstrate that the DA-SPS model outperforms existing state-of-the-art methods. Additionally, its performance in real-world scenarios is further validated using a private dataset collected by ourselves, which contains the test items' information on laptop motherboards.

\end{abstract}


\section{Introduction}
\IEEEPARstart{M}ultivariate time-series forecasting (MTSF) refers to using the historical information of variables to predict the future state value of variables, and it has a wide range of practical application scenarios, such as network traffic prediction \cite{10568345}, wind speed forecasting \cite{7091914}, and traffic prediction \cite{10680338, 9618828}. It can help us to make a response strategy in advance for possible situations in the future. For instance, the traffic control centre can prepare for evacuation in advance based on the traffic flow predicted by the system. MTSF is categorised into two types: multivariate input with multivariate output, and multivariate input with univariate output. This study focuses on the latter. We refer to the output variable as the target variable, while the other variables are called extraneous variables.

Traditional prediction models based on mathematical statistics, such as the autoregressive integrated moving average (ARIMA), are simple and effective. However, they often require the time series to be stationary and can only effectively capture linear features. Moreover, there are often multiple variables in the actual scenario while the mathematical statistical models cannot capture the complex nonlinear relationship between the variables. With the emergence and development of deep learning, this problem can be solved to a certain degree.

Due to its excellent nonlinear fitting ability, deep learning models can have a far-reaching impact in the field of MTSF, promoting the progress of MTSF research. Recurrent neural network (RNN) \cite{lai2018modeling}, a commonly used deep learning model, allows it to memorize historical information due to its cyclic structure. This feature makes it effective in extracting the temporary information of the input sequence. However, due to the gradient explosion problem of RNN, the deep learning models based on RNN cannot handle long input sequences well. To better remember long-term information, the long short-term memory network (LSTM) \cite{schmidhuber1997long, Fu2022, xiao2021dual}, a variant of RNN, effectively controls what information needs to be remembered or should be forgotten at each moment by introducing a gating mechanism, which effectively alleviates the problem of gradient explosion caused by the long input sequence. With the development of convolutional neural networks (CNN) in computer vision, more and more studies have tried to apply CNN to time series prediction \cite{ma2023tcln, xiao2021dual}. CNN can extract more temporal features because it can adjust the receptive field by changing the size of the convolutional kernel, which is beneficial to extracting seasonality features. To capture the relationships among global variables, the attention mechanism \cite{Qin-DA-RNN, Wu2021, ma2024fmamba} is widely used and boosts the models' performance. In this paper, MTSF tasks are mainly realised based on CNN, LSTM, and attention mechanism.

Currently, in MTSF tasks, the correlations between extraneous variables and the target vary significantly depending on the specific prediction target \cite{nguyen2024learning}. Therefore, the effects of extraneous variables on the prediction target are different. For instance, taking into account the variables that exhibit strong correlations with the target variate can enhance the model's predictive performance. However, variables that are weakly correlated or unrelated to the target variate can lead to the model learning unnecessary or even detrimental information, thereby introducing bias in the model's predictions. In addition, when analyzing variable sequences, it is evident that they generally contain different temporal patterns, such as trend, seasonality, and noise components\cite{ghil2002advanced}. Thus, when conducting analysis, it is important to consider these distinct temporal characteristics. For instance, trend and seasonality components can be effectively extracted by the model due to their stability and regularity, while noise components are often unpredictable, as they represent random fluctuations and errors in the data. However, current deep learning-based methods either fail to fully consider the impact of extraneous variables on the prediction target \cite{Qin-DA-RNN, lai2018modeling, Fu2022, ma2023tcln}, or typically extract features from complex variable sequences in a direct manner, without fully leveraging the information specific to each temporal pattern within the sequence\cite{yi2024deep, yang2024wind}. And the lack of sufficient consideration of these two factors always affects the model's prediction performance.

Therefore, to solve the above drawbacks and enhance the model's generalisation ability, a \textbf{d}ual-st\textbf{a}ge prediction model based on \textbf{S}ingular Spectrum Analysis (SSA), \textbf{P}atching-strategy and \textbf{S}pearman-correlation (DA-SPS) is proposed. DA-SPS employs a dual-stage model structure, the target variable processing stage (TVPS) and the extraneous variables processing stage (EVPS). 
Specifically, the TVPS uses SSA to decompose the target variable sequence into trend, seasonality, and noise components, where the noise component is considered interference and thus discarded. The trend and seasonality components are processed by the LSTM and P-Conv-LSTM modules, respectively, to extract the corresponding pattern features.
The EVPS initially uses the Spearman correlation algorithm to compute the correlation coefficient between the target variable and extraneous variables. It then eliminates variables that have irrelevant or weak correlations with the target variable and subsequently calculates the temporal dependencies among the retained variables through the proposed L-Attention module. Finally, the features obtained from TVPS and EVPS are combined through weighted summation and then passed through a linear layer to produce the final predicted values.
The principal contributions of this paper can be summarized as follows:
\begin{itemize}
    \item We introduce a novel prediction model, named DA-SPS, and experimental results indicate that DA-SPS surpasses the current state-of-the-art methods.
    \item We decompose the complex sequence into a series of local subsequences by SSA theory and patching strategy, which can effectively extract the temporal features of the target variable.
    \item We design a dual-stage network to extract the temporal features of the target variable and the correlation features between extraneous variables and the target variable separately. This design can fully capture the temporal patterns of the target variable and consider the impact of extraneous variables on the prediction of the target variable.
\end{itemize}

The rest of this paper is as follows: Section \ref{related work} mainly introduces the work related to MTSF; Section \ref{definition} mainly provides the MSTF problem definition; Section \ref{method} mainly illustrates the composition structure of the DA-SPS. Section \ref{experiments} mainly compares the performance of DA-SPS and the existing models and verifies the rationality of the model's design through ablation experiments. Finally, Section \ref{conclusion} summarizes the full text.

\section{Related Work} \label{related work}

Before the rise of deep learning methods, mathematical and statistical models like autoregressive \cite{gan2014seasonal} and ARIMA models \cite{mondal2014study, xiao2022research} were commonly employed for time series prediction tasks. For instance, Gan et al. \cite{gan2014seasonal} used a quasi-linear autoregressive model to predict the seasonal and trend time series. Mondal et al. \cite{mondal2014study} deployed the ARIMA model to forecast stocks. However, they struggled to handle the intricate nonlinear relationships both within and between variables. With the emergence and advancement of deep learning, there has been a growing interest in applying it to time series prediction tasks. Traditional deep learning models primarily relied on RNN, which could face gradient vanishing or explosion when processing long sequences. To address these challenges, variants of RNN such as GRU \cite{9749946} and LSTM were introduced. These variants aimed to mitigate gradient-related problems, enhancing the capability of capturing long-term features.

In addition, CNN is widely employed in time series prediction tasks. Lai et al. \cite{lai2018modeling} introduced LSTNet, which initially extracts short-term local dependencies between variables using convolutional layers, followed by capturing long-term patterns in the input series through RNN layers. Bai et al. \cite{bai2018empirical} proposed a novel temporal convolutional network based on CNN, enhancing the network's receptive field and aiding in capturing local temporal dependencies. Xiao et al. \cite{xiao2021dual} transformed the one-dimensional sequence of each moment into two-dimensional data, utilizing convolutional layers to extract features from this data, thereby capturing spatial dependencies between variables. Ma et al. \cite{ma2023tcln} introduced a multi-kernel CNN layer enabling the model to focus on local dependencies across different scales. However, despite considering the influence of extraneous variables on the target variable, these models overlook the impact of irrelevant information between variables on prediction accuracy. In contrast, the two-stage model proposed in this paper not only considers the correlation between extraneous variables and the target variable but also mitigates the influence of irrelevant information on prediction accuracy.

Time series data is inherently complex, and directly extracting features from the input series may not facilitate effective model learning. To enhance the model's ability to capture the predictable temporal features, the concept of time series decomposition is applied to prediction models \cite{ma2024multivariate, Wu2021}. For instance, Wu et al. \cite{Wu2021} introduced Autoformer, which decomposes input sequences into various subsequences using a temporal decomposition module, such as the trend and seasonality components. Similarly, Sulandari et al. \cite{sulandari2020time} proposed SSA-LRF, combining SSA with linear recurrent formulas to isolate signal components from noise in the series. Venkateswaran et al. proposed SSA-CNN-LSTM \cite{venkateswaran2024efficient}, combining SSA, CNN, and LSTM, to predict solar power generation. Moreno et al. \cite{moreno2024enhancing} explored two decomposition methods: SSA within the time domain and variational mode decomposition within the frequency domain, integrated into a hybrid framework that features multi-stage decomposition along with time series forecasting. However, while these models excel at extracting features from different temporal patterns, they often overlook dependencies between local information \cite{Yuqietal-2023-PatchTST}. In contrast, our approach combines SSA with the patching strategy to capture dependencies among local information across various time series patterns, particularly focusing on the trend pattern.

\begin{figure*}[ht]
	\centering
	\includegraphics[width=0.85\linewidth]{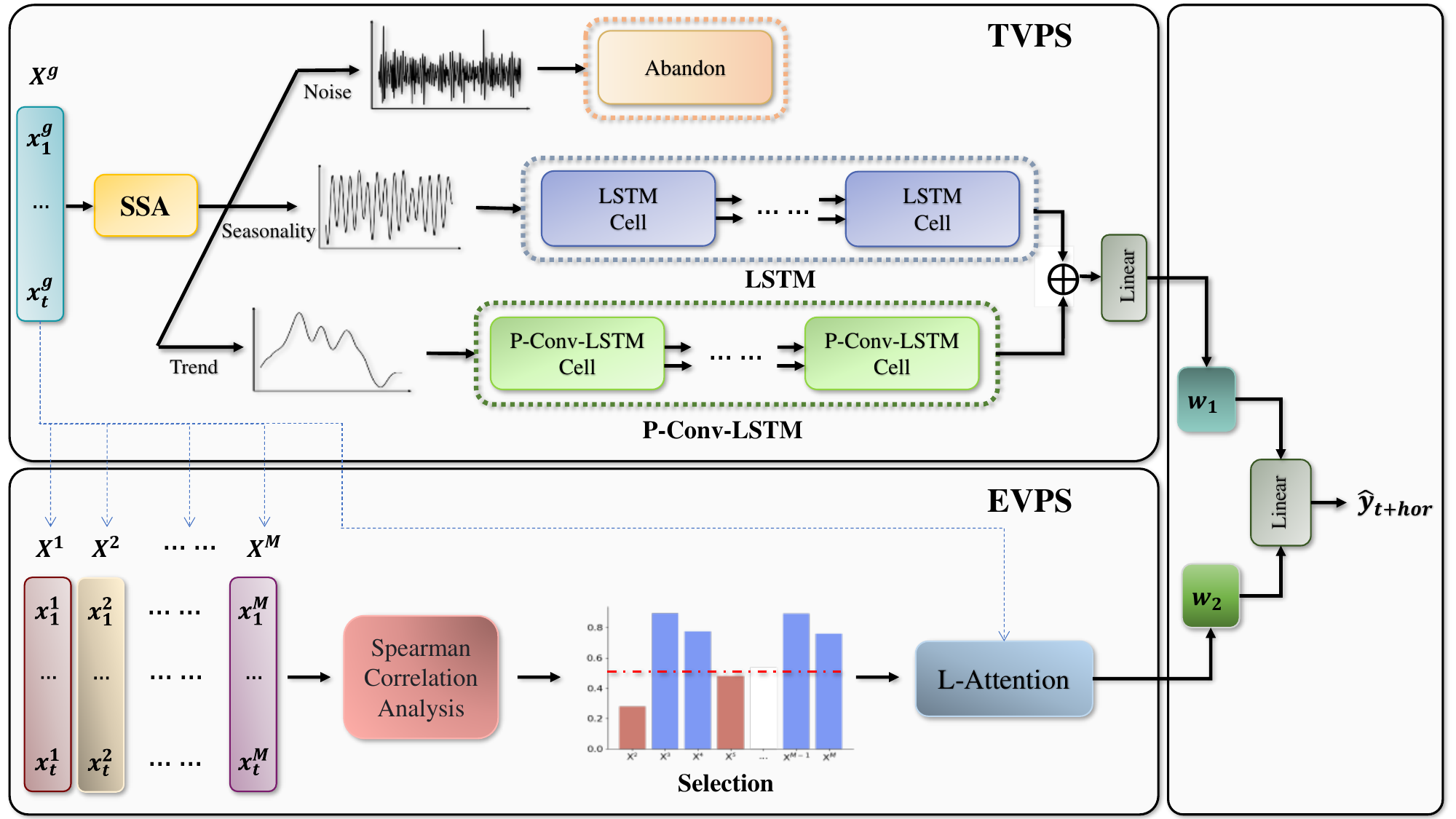}
	\caption{The framework of the DA-SPS model.}
	\label{fig:model_show_SSMa}
\end{figure*}

To calculate the global correlation between different time steps and variables and enhance the interpretability of the model\cite{xie2022deep}, the attention mechanism is applied to many prediction models. Qin et al. proposed DA-RNN \cite{Qin-DA-RNN}, which adaptively extracts relevant input features by introducing an input attention mechanism in the first stage, and selects the encoder hidden state at all time steps in the second stage by using the temporal attention mechanism. Fu et al. proposed a new self-attention mechanism that uses LSTM networks instead of simple matrix multiplication to encode input sequences, thereby retaining the temporal features \cite{Fu2022}. Ma et al. proposed the PCDformer \cite{ma2024multivariate}, which incorporated the sparse attention and alleviated the impact of the useless features. Jiang et al. \cite{jiang2023fecam} introduced an innovative frequency-enhanced channel attention mechanism leveraging Discrete Cosine Transform to capture frequency interdependencies across channels, which enhances the model's ability to extract frequency features while addressing computational overhead issues associated with inverse transformations.

\section{Preliminary} \label{definition}

In this section, we provide a definition of the MSTF problem. Consider a dataset denoted as $X_{\operatorname{dataset}} = (X_1,X_2, \dots,X_N)^T$, where $N$ represents the length of the dataset. $X_t = (x_{t}^1,x_{t}^2,\dots,x_{t}^M) \in \mathbb{R}^M$ contains the values of all $M$ variables at time $t$. $M$ denotes the total number of variables of the dataset, and the target variable is denoted as $y = X^g$, with $g \in [1, M]$.

The primary objective of this study is to forecast the target variable's value at various future time steps, denoted as the horizon, which could be 3, 6, 12, or 24. The prediction task involves analyzing the sequence values of the past $t$ time steps and making predictions toward various horizon values. For instance, given input data $X = (X_1, X_2, \dots, X_t)^T $ and a horizon $hor$, the forecasted value of the target variable would be denoted as $\hat y_{t+hor}$. Building upon the above interpretation, our goal can be concisely formulated as follows:
\begin{equation}
	\hat y_{t+hor} = f\left(X_1, X_2, \dots,X_t, \text{horizon}=hor \right) ,
\end{equation}
where $f\left(\cdot \right)$ denotes the mapping function that the DA-SPS needs to learn.

\section{Methodology} \label{method}

This section begins with a summary of the main structure of the model and then introduces the details of the model in two parts: TVPS and EVPS.

\subsection{Model Architecture}

The overall architecture of DA-SPS is shown in Fig.~\ref{fig:model_show_SSMa}. In the model, TVPS and EVPS are respectively designed for processing the target variable sequence and extraneous variable sequences.

\begin{figure*}[ht]
	\centering
	\includegraphics[width=0.85\linewidth]{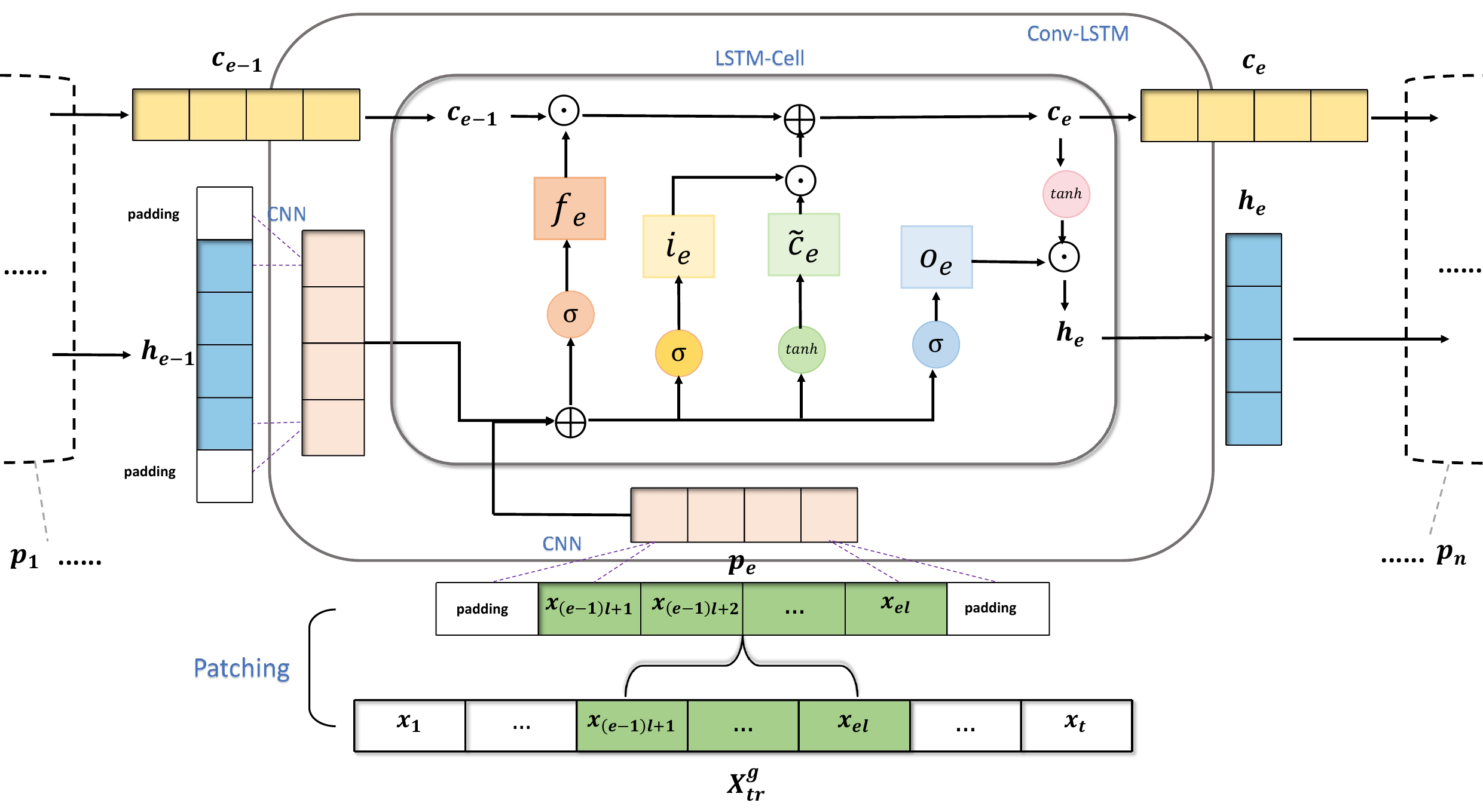}
	\caption{The structure of the P-Conv-LSTM.}
	\label{fig:P-Conv_lstm_SSMa}
\end{figure*}

In the TVPS, we initially use the sequence decomposition algorithm to decompose the target variable sequence into the trend, seasonality, and noise components. After removing the noise component, two distinct modules, LSTM and P-Conv-LSTM, are utilized to extract the features of the seasonality and trend components, respectively. The structure of P-Conv-LSTM is shown in Fig.~\ref{fig:P-Conv_lstm_SSMa}. In the final of the TVPS, we integrate and input the features extracted by both modules into a linear layer to fully learn diverse temporal patterns of the target variable sequence.

\begin{figure*}[ht]
	\centering
		\includegraphics[width=0.85\linewidth]{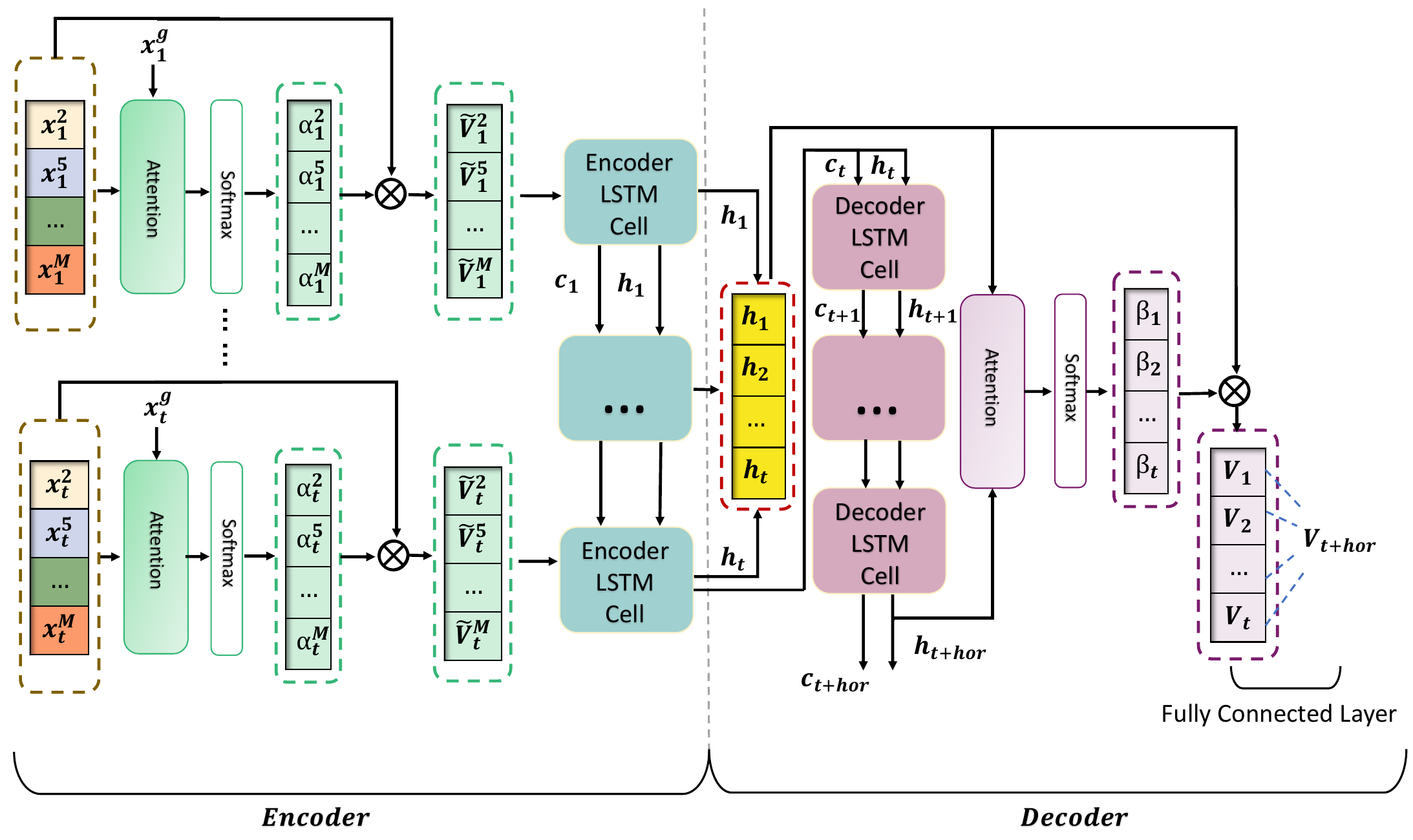}
	\caption{The structure of the L-Attention. Here, we assume that the selected variables are variables 2, 5, $\dots$, and $M$. The total number of the variables is $Q$.}
	\label{fig:L-Attention-ZTH}
\end{figure*}

For the extraneous variables' sequences, the EVPS first identifies those that have strong correlations with the target variable sequence by computing and comparing the Spearman Correlation Coefficient for each sequence. Then, the related variables' sequences are selected as inputs to the L-Attention module, whose architecture is shown in Fig.~\ref{fig:L-Attention-ZTH}. 

Finally, the features obtained from the TVPS and EVPS are combined through a fully connected layer to predict the target variable sequence's future value at $hor_{th}$ time step.

\subsection{TVPS}

In this stage, we conduct sequence decomposition by SSA and primarily focus on predictable seasonality and trend components. For the seasonality component, we use the LSTM module to extract seasonality features. As for the more intricate trend component, the proposed P-Conv-LSTM module utilizes the patching strategy to segment it, followed by the Conv-LSTM network to capture temporal dependencies.

\subsubsection{SSA Module}

In the TVPS, the SSA method was selected for the decomposition of the target variable sequence. SSA has proven to be an effective methodology in dealing with nonlinear sequences \cite{venkateswaran2024efficient, moreno2024enhancing}, which contains four phases: embedding, Singular Value Decomposition (SVD), grouping, and reconstruction. By this method, the target variable sequence can be decomposed into the trend sequence $(X^{g}_{tr})$, seasonality sequence $(X^{g}_{se})$, and noise sequence $(X^{g}_{no})$. The following is about the details of the decomposition:

First, construct a trajectory matrix $\Ddot{X}^{g}$ of dimension $(m,l)$. For the input target sequence $X^g = {\{x^g_1,x^g_2,x^g_3,...x^g_t\} \in\mathbb{R}^t}$, the $\Ddot{X}^{g}$ can be expressed as:
\begin{equation}\label{eq:SVD_matrix_THZhang}
    \Ddot{X}^{g} =  \begin{bmatrix}
            x^g_1&x^g_2&...&x^g_l\\
            x^g_2&x^g_3&...&x^g_{l+1}\\
            ...&...&...&...\\
            x^g_m&x^g_{m+1}&...&x^g_t\\
            \end{bmatrix} ,
\end{equation}
where $m$ fulfills the condition $1<m<t$ and $l$ meets the requirement that $l=t-m+1$.

Next we perform SVD to $\Ddot{X}^{g}$, which can be represented as $\Ddot{X}^{g}=DH{E^{\mathrm{T}}}$. $D$ and $E$ denote the left and right singular matrices of $\Ddot{X}^{g}$, respectively. $H$ is a diagonal matrix composed of singular values.
Define the covariance matrix $S={{\Ddot{X}^{g}}}({\Ddot{X}^{g}})^{\mathrm{T}}$, and then $S$ can be represented as $S=D{H^2}{D^{\mathrm{T}}}$. $H^2$ is a diagonal matrix consisting of $m$ characteristic values, where the characteristic values satisfy $\{\lambda_1 \geq \lambda_2 \geq ... \geq \lambda_m \geq 0\}$, and each characteristic value $\lambda_i$ corresponds to a pair of orthogonal vectors $D_i$ and $E^{\mathrm{T}}_i$. Thus, the $\Ddot{X}^{g}$ can be expressed by the Eq.~\eqref{eq:SVD_decompose_THZhang}:
\begin{equation}\label{eq:SVD_decompose_THZhang}
    \Ddot{X}^{g}=\Ddot{X}^{g}_{1}+\Ddot{X}^{g}_{2}+...+\Ddot{X}^{g}_{d} ,
\end{equation}
where $d=\text{rank}(\Ddot{X}^{g})=\max(i)(\lambda_i>0)$, $\Ddot{X}^{g}_{i}=\sqrt{\lambda_i} D_i E^{\mathrm{T}}_i$,  and $E_i=\frac{({\Ddot{X}^{g}})^{\mathrm{T}} D_i}{\sqrt{\lambda_i}}$.

During the grouping phase, we reclassify the set of subscripts $\{1, 2, 3, ..., d\}$ into $3$ mutually exclusive subsets $I=\{I_{tr}, I_{se}, I_{no}\}$. Thereby, the trajectory matrix $\Ddot{X}^{g}$ will be divided into $3$ groups:
\begin{equation}\label{eq:SVD_grouping_THZhang}
    \Ddot{X}^{g} = \Ddot{X}^{g}_{{I_{tr}}} + \Ddot{X}^{g}_{{I_{se}}} +  \Ddot{X}^{g}_{{I_{no}}}
\end{equation}

In the final phase, the grouped $\Ddot{X}^{g}_{{I_f}}$ will be reconstructed into a sequence $X^{g}_{f}$ of length $t$ by using diagonal averaging, where $f \in \{tr, se, no\}$. The formula is as follows:
\begin{equation}\label{eq:SVD_Reconstruct_THZhang}
    (x^{g}_{f})_i=
        \begin{cases}
            \frac{1}{i} $$\sum_{j=0}^i$$ y_{(j,i-j+1)}^* 
            &\text{$1 \leq i < m^*$}\\
            \frac{1}{m^*}$$\sum_{j=1}^{m^*} $$y_{(j,i-j+1)}^*
            &\text{$m^* \leq i \leq  l^*$}\\
            \frac{1}{N-i+1}$$\sum_{j=i-l^*+1}^{N-l^*+1}$$y_{(j,i-j+1)}^*
            &\text{$l^* < i \leq  t$} ,\\
        \end{cases}
\end{equation}
where $m^*=min(m,l)$, $l^*=max(m,l)$, and $N=m+l-1$. And the $y_{ij}^*$ is satisfied:
\begin{equation}\label{eq:y_{ij}^*_THZhang}
    y_{ij}^* = 
        \begin{cases}
        \Ddot{X}^{g}_{{I_f}(ij)}  
        &\text{$m < l$}\\
        \Ddot{X}^{g}_{{I_f}(ji)}
        &\text{$m \geq l$}  .\\
        \end{cases}
\end{equation}

To ensure that the learned features are predictable, we only consider the decomposed $X^{g}_{se}$ and $X^{g}_{tr}$, eliminating the impact of the $X^{g}_{no}$ on the model. In the following, we will introduce separately how to extract the features from $X^{g}_{se}$ and $X^{g}_{tr}$.

\subsubsection{LSTM Module}

LSTM is a type of RNN that is widely used to capture the long-term dependencies of sequences. Considering the stable periodic changes throughout the whole seasonality component, LSTM can effectively learn and remember the long-term repeated features of its hidden units. Therefore, for the reconstructed $X^{g}_{se}$, we deploy the LSTM module for analysis. The primary formulas involved are as follows:
\begin{equation}\label{lstm_eq_THZhang}
    \begin{aligned}
    &f^{\prime}_t = \sigma({W^{\prime}_f}{[h^{\prime}_{t-1};x_t]}) + b^{\prime}_f\\
    &i^{\prime}_t = \sigma({W^{\prime}_i}{[h^{\prime}_{t-1};x_t]}) + b^{\prime}_i\\
    &o^{\prime}_t = \sigma({W^{\prime}_o}{[h^{\prime}_{t-1};x_t]}) + b^{\prime}_o\\
    &\Tilde{c}^{\prime}_t=f^{\prime}_t{\circ \,}{c^{\prime}_{t-1}} + i^{\prime}_t{\circ \,}{tanh}({W^{\prime}_{\Tilde{c}}{[h^{\prime}_{t-1};x_t]}+b^{\prime}_{\Tilde{c}}})\\
    &h^{\prime}_t=o^{\prime}_t{\circ \,}{tanh}({\Tilde{c}^{\prime}_t}) ,\\
    \end{aligned} 
\end{equation}
where ${x_t}$ is the value of $X^{g}_{se}$ at the $t$ moment, and $h^{\prime}_{t-1}$ corresponds to the hidden state from the previous time step. 
The $\sigma$ stands for the activation function as shown in Eq.~\eqref{sigmoid_eq_THZhang} and the symbol ${\circ \,}$ represents the Hadamard operation.
${f^{\prime}_t,i^{\prime}_t,o^{\prime}_t, \Tilde{c}^{\prime}}$  respectively represent the forgetting gate, input gate, output gate, and memory cell candidate of LSTM. 
The ${W^{\prime}_f},{W^{\prime}_i},{W^{\prime}_o},{W^{\prime}_{\Tilde{c}}},b^{\prime}_f,b^{\prime}_i,b^{\prime}_o,$ and $b^{\prime}_{\Tilde{c}}$ denote trainable parameters.

\begin{equation}\label{sigmoid_eq_THZhang}
    \sigma(x) = \frac{1}{1+e^{-x}} .
\end{equation}

\subsubsection{P-Conv-LSTM Module}

In the traditional LSTM network, the input data acquired by each cell is the data of a single-time step. 
While this design can extract temporal dependence between different steps, it ignores the connection between local information in the sequence and too many cells can also lead to the problem of gradient vanishing or exploding, which affects the model training.

To fully exploit the local information of the $X^{g}_{tr}$ and avoid the effect of too many cells, we propose the P-Conv-LSTM module.
This module initially uses the patching strategy to segment $X^{g}_{tr}$ into non-overlapping subsequences and then captures the local features and analyzes them through the Conv-LSTM network.

Assuming the length of each subsequence is $l$, the patching strategy can be expressed as follows:
\begin{equation}\label{eq:Patch_1_THZhang}
    \begin{aligned}
       & \operatorname{Patching}(X^{g}_{tr})=\{p_1, p_2, p_3, ..., p_e, ..., p_n\} \\
    & p_e=\{x_{(e-1)l+1},x_{(e-1)l+2},...,x_{el}\} ,
    \end{aligned}
\end{equation}
where n denotes the number of subsequences and $n=\lfloor {\frac{t}{l}}\rfloor$. The patching structure makes it easier to capture the local features by converting the long sequence into a series of simple subsequences.

Subsequently, considering the limitations of fully connected LSTM (FC-LSTM) in handling long-term high-dimension data, we use the Conv-LSTM structure \cite{shi2015convolutional} to better capture the subsequences' local patterns and structural information.
The relevant formulas for this structure are as follows:
\begin{equation}\label{conv_lstm_eq_THZhang}
    \begin{aligned}
    &i_e=\sigma({W_{p}^i}*p_e + {W_{h}^i}*h_{e-1} + W_{c}^i{\circ \,}{C_{e-1}} + b^i)\\
    &f_e=\sigma({W_{p}^f}*p_e + {W_{h}^f}*h_{e-1} + W_{c}^f{\circ \,}{C_{e-1}} + b^f)\\
    &\Tilde{c}_e=f_e{\circ \,}{C_{e-1}} + i_e{\circ \,}{tanh}({W_{p}^{\Tilde{c}}}*p_e + {W_{h}^{\Tilde{c}}}*h_{e-1}+b^c)\\
    &o_e=\sigma({W_{p}^o}*p_e + {W_{h}^o}*h_{e-1} + W_{c}^o{\circ \,}{\Tilde{c}} + b^o)\\
    &h_e=o_e{\circ \,}{tanh}({\Tilde{c}}) , 
    \end{aligned}
\end{equation}
where the $p_e{\in}{\mathbb{R}}^{P\times1}$ denotes the input of the $e$-th Conv-LSTM unit and the $h_{e-1}{\in}{\mathbb{R}}^{P\times1}$ expresses the hidden state from the previous unit. 

The symbol $*$ denotes one-dimensional convolution operation, the ${i_t, f_t,\Tilde{c}, o_t}$ respectively represent the forgetting gate, input gate, output gate, and memory cell candidate in each cell, and the ${W_p, W_h, b}$ are the learnable parameters.

In summary, the P-Conv-LSTM module uses the patching strategy to segment the trend sequence and then deploys the Conv-LSTM structure to extract local features within the subsequence. This design not only enables the model to better handle long-term sequence but also allows for a more detailed local analysis.

\subsection{EVPS}

For a precise prediction of the target variable, it is essential to take into account the influence of extraneous variables aside from considering the historical data of the target variable itself.
When extraneous variables' sequences have a strong temporal correlation with the target variable sequence, the input of these variables' sequences will facilitate the model to better learn the temporal characteristics of the target variable sequence.
Conversely, the input of irrelevant variables' sequences will interfere with the model's learning and inference toward the target variable sequence.

Therefore, for the MTSF task with multivariate input and univariate output, we need to identify and select the extraneous variables that exhibit strong correlations with the target variable.
These carefully related variables can then be employed to aid the model in learning the patterns of the target variable more effectively.

\subsubsection{{Spearman Correlation Analysis} Module}

Considering the complex correlations between variables in real-world data, the strength of the correlation between each variable and the target variable varies. To minimize the influence of irrelevant or weakly relevant variables on the prediction task, filtering the irrelevant variables is necessary.

In this paper, we choose the Spearman correlation analysis to filter the variables. The formula is shown as follows:
\begin{footnotesize}  
\begin{equation}\label{spearman-eq-thZhang}
    {\rho}=\frac{{\frac{1}{t}}{{\sum_{n=1}^{t}}((R(x^g_n)-\overline{R(X^g)})\cdot(R(x^k_n)-\overline{R(X^k)}))}}{\sqrt{({\sum_{n=1}^{t}}({(R(x^g_n)-\overline{R(X^g)})^2})\cdot({\sum_{n=1}^{t}}({(R(x^k_n)-\overline{R(X^k)})^2})}} .
\end{equation}
\end{footnotesize}
${x^g_n}$ represents the value of the target variable, ${X^g}$, at the moment of ${n}$. ${x^k_n}$ denotes the value of the extraneous variable, ${X^k}$, at the moment of ${n}$.
$R(x^g_n)$ and $R(x^k_n)$ represent the position in ${X^g}$, ${X^k}$ by size. $\overline{R(X^g)}$ and $\overline{R(X^k)}$ represent the average position. ${\rho}$ denotes the Spearman Correlation Coefficient between ${X^k}$ and ${X^g}$.

The absolute value of ${\rho}$ lies between 0 and 1, with larger values indicating a stronger correlation. 
By obtaining the coefficients, we can infer the relative strength of the correlation between extraneous variables and the target variable. A higher coefficient implies a closer and more obvious relationship. Subsequently, each of the related variable sequence ${X^z}$, ${z\in\{2,5,..., M\}}$, is filtered by setting a threshold $\rho^{\prime}$. Following this, the L-Attention module is deployed for further analysis.

\subsubsection{L-Attention Module}

Following the previous analysis between the extraneous variables and the target variable, we propose the L-Attention based on LSTM and dual-attention to more effectively extract and use the correlation information between ${X^z}$, ${z\in\{2,5,..., M\}}$ and the target variable ${X^g}$ for better prediction.

The L-Attention module consists of two parts: an encoder and a decoder. The encoder mainly contains two structures, the attention mechanism and LSTM, where the attention mechanism is designed to capture the correlation among variables.

For the input at time $t$, denoted as $\{x^2_t,x^5_t,...,x^M_t\} \in {\mathbb{R}^{Q}}$, the Attention layer is constructed using a Multi-Layer Perceptron (MLP)\cite{Qin-DA-RNN}, with the following formula:
\begin{equation}\label{encoder_eq_THZhang}
        e^z_t = {v_e^T} {tanh} {(W^z x^{z}_{t} + b^z x_{t}^{g})},
\end{equation}
where ${v_e}$, $W^z$, and ${b^z}\in{\mathbb{R}^{Q \times 1}}$ are the learnable parameters. 
We then use the Softmax function to obtain the weight ${\alpha^z_t}$. The formula is shown in Eq.~\eqref{edcoder_softmax_THZhang}:
\begin{equation}\label{edcoder_softmax_THZhang}
    \alpha^z_t = \frac{exp(e^z_t)}{\sum_{k=2}^{M}exp(e^k_t)}, \quad k \in \{2, 5, ..., M\}.
\end{equation}

Subsequently, the set of attention weights, ${\{\alpha^2_t, \alpha^5_t,...,\alpha^M_t\}}$, is multiplied with their corresponding feature inputs to obtain new input, ${\{\Tilde{V}^2_t, \Tilde{V}^5_t,...,\Tilde{V}^M_t\}}$, which incorporate the attention information. The new input is fed into the LSTM network to learn the weights and temporal information. Therefore, the encoder output is represented as shown in Eq.~\eqref{encoder_output_eq_THZhang}:
\begin{equation}\label{encoder_output_eq_THZhang}
    \begin{aligned}
    &h_{1:t}= \operatorname{EncoderLSTM}[\Tilde{V}_{1:t}],\\
    \end{aligned}
\end{equation}
where $\operatorname{EncoderLSTM(\cdot)}$, similar to Eq.~\eqref{lstm_eq_THZhang}, denotes the encoder LSTM network, where the number of cells is determined by the total steps $t$ of the sequence. $h_{1:t}$ denotes the collection of outputs from the encoder, and $\Tilde{V}_{1:t}$ represents the set of all inputs from time $1$ to $t$, where each input has been weighted accordingly.

A decoder corresponding to the encoder is designed to excavate relevant feature information further and enhance prediction accuracy in the long-term prediction task. The decoder can be expressed as follows:
\begin{equation}\label{decoder_eq_THZhang}
    \begin{aligned}
    &V_{t+hor}= W_{d}[{\sum_{k=1}^{t}(\beta_{1:t}*h_{1:t})}]+b_{d}\\
    &\beta_{1:t} = \operatorname{AttnScore}[h_{1:t},h_{t+hor}]\\
    &h_{t+hor} = \operatorname{DecoderLSTM}[c_{t}, h_{t}],
    \end{aligned}
\end{equation}
where ${hor}$ represents the horizon, and $\beta_{1:t}$ represents the attention weights from the $h_{1:t}$. $W_{d}$ and $b_{d}$ are the learnable parameters. $\operatorname{AttnScore(\cdot)}$ is similar to Eq.~\eqref{encoder_eq_THZhang} and \eqref{edcoder_softmax_THZhang}. $h_{t+hor}$ represents the hidden state at the ${hor}$ time step of the decoder LSTM layer.
The $\operatorname{DecoderLSTM}(\cdot)$ is similar to Eq.~\eqref{lstm_eq_THZhang}, where $c_{t}$ and $h_{t}$ represent the cell state and hidden state, respectively, at the $t$ time step of the encoder LSTM layer.

In the decoder, the outputs obtained from the encoder are first fed into the decoder LSTM cell for decoding, where the number of cells in the LSTM depends on the size of the prediction horizon ${hor}$. After that, the decoder attention mechanism is used to derive the weights of the encoder layer's outputs $h_{1:t}$ to $h_{t+hor}$. Next, the Softmax function is applied to obtain the normalised weight set ${\{\beta_1, \beta_2,...,\beta_t\}}$, which is then multiplied with the $h_{1:t}$ to get the output ${\{V_1, V_2,..., V_t\}}$. Finally, a fully connected layer is used to obtain the EVPS output $V_{t+hor}$.

Overall, the design of the encoder in L-Attention allows the model to gather data information from both the temporal features within the sequence itself and the correlations between variables. This can enhance the effective use of information from related variables, improving the prediction accuracy of the target variable. 

At the end of the model, the outputs from TVPS and EVPS are combined through weighted summation and a linear mapping to get the predicted result, $\hat y_{t+hor}$.

\section{Experiments}  \label{experiments}

In this section, we first introduce four public datasets adopted for validating the performance of the model. Then, we describe the metrics used to evaluate the model's performance. Next, we illustrate the experimental details. Further, we specifically introduce the Laptop Board Yield (LBY) dataset and demonstrate the validity of the DA-SPS model on this dataset. Finally, through the ablation studies, we substantiate the importance and effectiveness of each module within the DA-SPS. We also compare and analyze two sequence decomposition algorithms to reveal their respective strengths and weaknesses. All experiments in this paper are executed on NVIDIA GeForce RTX 3060, AMD Ryzen 7 and 16GB of RAM.

\subsection{Dataset Introduction}\label{sec:data_introduction}

We choose Electricity, Solar, Traffic, and Exchange datasets to validate the model effect. The specific details of the four datasets are shown in TABLE~\ref{tab:data_descrip_THZhang}. The relevant information about the datasets is as follows:

\textbf{Electricity}: The electricity consumption in kWh was recorded every 15 minutes from 2012 to 2014, for n=321 clients. We converted the data to reflect hourly consumption.

\textbf{Solar}: The solar power production records in 2006, which are sampled at 10-minute intervals from 137 PV plants in Alabama state. 

\textbf{Traffic}: The California Department of Transportation data cover a total of 24 months, from January 2015 to December 2016, describing road occupancy rates (ranging from 0 to 1) measured by various sensors on freeways in the San Francisco Bay area.

\textbf{Exchange}: Contain daily exchange rates from 1990 through 2016 for eight countries: Australia, the United Kingdom, Canada, Switzerland, China, Japan, New Zealand, and Singapore.

\begin{table}[htbp]
\centering
\fontsize{9pt}{9pt}\selectfont
\centering
\caption{Description of the datasets. $n_{th}$ denotes the $n_{th}$ variable of the dataset in the original order.}
\begin{tabular}{c|c|c|c}
\toprule [1.0pt]
Dataset &  Target &  Time series &  Variables\\
\midrule [1.0pt]
Electricity & $321_{th}$ & 320+1 & 26,304  \\
Solar       & $1_{th}$   & 136+1 & 52,560  \\
Traffic     & $862_{th}$ & 861+1 & 17,544  \\
Exchange    & $1_{th}$   & 7+1   & 7,588 \\
\bottomrule [1.0pt]  
\end{tabular}
\label{tab:data_descrip_THZhang}
\end{table}

All the above datasets are divided into training set  ($60\%$), validation set ($20\%$), and testing set ($20\%$) to facilitate the model's training, validating, and testing processes.

\subsection{Metrics}\label{sec:Metrics}

To comprehensively evaluate the model performance, we use four metrics: Mean Absolute Error (MAE), Root Mean Square Error (RMSE), Relative Squared Error (RSE), and Empirical Correlation Coefficient (CORR) to measure the model prediction results. Among these four metrics, for MAE, RMSE, and RSE, lower values indicate higher predictive accuracy. In contrast, a higher value for CORR suggests a stronger correlation between the model's predicted and true values. The formula for each metric is as follows:
\begin{small} 
\begin{align}
    \begin{split}
    &\operatorname{MAE} = \frac{1}{n} \sum_{t=1}^{n} \lvert y_t-\widehat{y}_t \rvert \\
    &\operatorname{RMSE} = \sqrt{\frac{1}{n} \sum_{t=1}^{n} (y_t-\widehat{y}_t)^2} \\
    &\operatorname{RSE} = \sqrt{\frac{{\sum_{t=1}^{n}(y_t-\widehat{y}_t)^2}}{{\sum_{t=1}^{n} (y_t-{\operatorname{mean}}({y}))^2}}} \\
    &\operatorname{CORR}=\frac{1}{n} \frac{\sum_{t=1}^{n}{(y_t-{\operatorname{mean}}({y}))(\widehat{y}_t-{\operatorname{mean}}({\widehat{y}}))}}{\sqrt{\sum_{t=1}^{n}({y_t-{\operatorname{mean}}({y})})^2 ({\widehat{y}_t-{\operatorname{mean}}({\widehat{y}}))^2}}} .
    \end{split}
\end{align}
\end{small}
$y_t$ and $\widehat{y}_t$ represent the true and predicted values at time step $t$, respectively. $n$ is the total length of the time steps. ${\operatorname{mean}}({y})$ and ${\operatorname{mean}}({\widehat{y}})$  respectively denote the mean values of the true and predicted values.

\subsection{Experimental Details}\label{sec:Experimental details}

We use MAE as the loss function and apply the Adam optimization algorithm to optimize during the training process with backpropagation to update the model parameters. 

To verify the performance of the DA-SPS, seven baselines are selected for comparison, namely CNN, LSTM, DA-RNN \cite{Qin-DA-RNN}, LSTNet \cite{lai2018modeling}, DA-Conv-LSTM \cite{xiao2021dual}, TS-Conv-LSTM \cite{Fu2022}, and TCLN \cite{ma2023tcln}.

\begin{table*}[htbp]
\centering
\fontsize{9pt}{9pt}\selectfont
\centering
\caption{The Results of All Models across Public Datasets for Various Tasks and Metrics: Optimal Values Shown in \textbf{Bold}, Sub-Optimal Values in \underline{Underline}.}
\label{tab:exp.mainResults}
\resizebox{\textwidth}{!}{
\begin{tabular}{c|c|c|c|c|c}
\toprule[1.0pt]
\multicolumn{2}{c|}{Datasets}  
& {Electricity}  & {Solar} & {Traffic} & {Exchange}  \\

\midrule[0.5pt]
\multicolumn{2}{c|}{Metric}          
& \textbf{3}\qquad~~\textbf{6}\qquad~~\textbf{12}\qquad~~\textbf{24}          
& \textbf{3}\qquad~~\textbf{6}\qquad~~\textbf{12}\qquad~~\textbf{24}           
& \textbf{3}\qquad~~\textbf{6}\qquad~~\textbf{12}\qquad~~\textbf{24}          
& \textbf{3}\qquad~~\textbf{6}\qquad~~\textbf{12}\qquad~~\textbf{24}      \\

\midrule[1.0pt]
\multirow{4}{*}{\rotatebox{90}{CNN}}	 
& 	MAE	 & 	
0.0357~~0.0352~~0.0410~~0.0406   & 	
0.0356~~0.0475~~0.0595~~\textbf{0.0587}  &
0.0322~~0.0365~~0.0346~~0.0349	 & 	
0.0146~~0.0214~~0.0261~~0.0299	 \\

&   RMSE	& 	
0.0464~~0.0466~~0.0547~~0.0542   & 	
0.0791 ~~\underline{0.0932}~~0.1158~~\underline{0.1243}	 & 	
0.0477~~0.0533~~0.0515~~0.0509   &
0.0171~~0.0290~~0.0315~~0.0350	 \\
  
&   RSE	& 	
0.4943~~0.4962~~0.5829~~0.5770   & 	
0.3184~~\underline{0.3749}~~0.4657~~\underline{0.5000}  &
0.4395~~0.4906~~0.4742~~0.4690   &
0.1719~~0.2924~~0.3172~~0.3520	 \\
  
&   CORR	 & 	
0.8703~~0.8714~~0.8464~~0.8488   & 	
0.9503~~0.9295~~0.8876~~\underline{0.8786}  &
0.9603~~0.9521~~0.9477~~0.9533  &
0.9932~~0.9792~~0.9722~~0.9696	 	 \\

\midrule[0.5pt]

\multirow{4}{*}{\rotatebox{90}{LSTM}}	 
&    MAE     & 	
0.0366~~0.0373~~0.0390~~\underline{0.0387}  & 	
\underline{0.0325}~~\textbf{0.0433}~~0.0588~~0.0680   & 
0.0361~~0.0369~~0.0388~~\underline{0.0306}	 & 	
0.0128~~0.0221~~0.0312~~0.0387		 \\

&    RMSE     & 	
0.0480~~0.0490~~0.0511~~0.0505    & 	
0.0749~~0.0970~~0.1211~~0.1415    &
0.0517~~0.0535~~0.0569~~\underline{0.0464}	 & 	
0.0151~~0.0252~~0.0351~~0.0440	 	\\
  
&    RSE     & 	
0.5108~~0.5215~~0.5437~~0.5381   & 
0.3013~~0.3903~~0.4886~~0.5692   &
0.4762~~0.4924~~0.5239~~\underline{0.4273}	 & 	
0.1518~~0.2533~~0.3531~~0.4428	 	\\
  
& 	 CORR	 & 	
0.8605~~0.8586~~0.8469~~0.8445   &
0.9549~~0.9269~~0.9044~~0.8636   &
\textbf{0.9662}~~\underline{0.9555}~~0.9502~~0.9559   & 	
\underline{0.9959}~~0.9916~~0.9855~~0.9713	 	\\

\midrule[0.5pt]

\multirow{4}{*}{\rotatebox{90}{DA-RNN}}	 
&    MAE     & 	
0.0324~~0.0426~~0.0446~~0.0440   &
0.0337~~0.0471~~\underline{0.0496}~~0.0611   &
0.0354~~0.0304~~0.0351~~0.0395	 &
0.0159~~0.0194~~0.0491~~0.0560	  \\

&    RMSE     & 	
0.0419~~0.0547~~0.0607~~0.0589   & 	
0.0729~~0.0944~~\underline{0.1043}~~0.1324	 & 	
0.0513~~0.0458~~0.0498~~0.0525	 &
0.0186~~0.0226~~0.0537~~0.0613	 	\\
  
&    RSE     &
0.4467~~0.5825~~0.6459~~0.6275    & 	
0.2934~~0.3798~~\underline{0.4196}~~0.5327	 &
0.4718~~0.4219~~0.4584~~0.4832	 & 	
0.1868~~0.2274~~0.5409~~0.6172	 	 \\
  
&    CORR     & 	
0.9004~~0.8254~~0.7942~~0.8675   &
0.9586~~0.9275~~\underline{0.9085}~~0.8664	 & 	0.9546~~0.9538~~0.9486~~0.9379	 & 	0.9951~~0.9922~~0.9756~~0.9729	 	 \\

\midrule[0.5pt]

\multirow{4}{*}{\rotatebox{90}{LSTNet}}	 
&    MAE     & 		
0.0311~~\underline{0.0335}~~\underline{0.0337}~~0.0405    &
0.0342~~0.0479~~0.0591~~0.0668	 & 
0.0290~~0.0342~~\underline{0.0312}~~0.0321	 &
0.0281~~0.0317~~0.0408~~0.0465	 	 \\

&    RMSE    & 		
0.0411~~\underline{0.0440}~~\underline{0.0452}~~0.0537   &
0.0727~~0.1014~~0.1135~~0.1405	 & 
0.0436~~0.0502~~0.0477~~0.0482	 & 	
0.0362~~0.0402~~0.0480~~0.0543	 	 \\
  
&    RSE     & 		0.4377~~\underline{0.4684}~~\underline{0.4809}~~0.5715    &
0.2927~~0.4081~~0.4565~~0.5655	 &
0.4014~~0.4625~~0.4390~~0.4434	 & 	
0.3644~~0.4049~~0.4832~~0.5462	 	   \\
  
&    CORR   & 	 	
0.9086~~\underline{0.8928}~~\underline{0.8896}~~0.8455    &
0.9579~~0.9170~~0.9073~~0.8671	 &
0.9543~~0.9538~~0.9457~~0.9385	 & 	
0.9831~~0.9726~~0.9607~~0.9537	 	   \\

\midrule[0.5pt]

\multirow{4}{*}{\rotatebox{90}{\fontsize{5.0pt}{\baselineskip}\selectfont DA-Conv-LSTM}}	 
&    MAE     & 		
\underline{0.0288}~~0.0382~~0.0439~~0.0406	     & 
0.0344~~0.0571~~0.0519~~0.0699		 & 	
\underline{0.0280}~~0.0385~~0.0422~~0.0332	     & 
0.0088~~0.0118~~0.0128~~0.0283	 \\

&    RMSE    &
\underline{0.0390}~~0.0495~~0.0583~~0.0564 	& 	
0.0718~~0.1109~~0.1044~~0.1345	 & 	
\textbf{0.0405}~~0.0547~~0.0575~~0.0474	 & 	
0.0113~~0.0146~~\underline{0.0160}~~0.0366	 \\
  
&    RSE    &
\underline{0.4153}~~0.5275~~0.6212~~0.6008	 & 	
0.2891~~0.4462~~0.4199~~0.5410	 & 	
\textbf{0.3727}~~0.5040~~0.5290~~0.4365	 &
0.1135~~0.1468~~\underline{0.1608}~~0.3682	 \\
  
&    CORR     & 	
\underline{0.9102}~~0.8796~~0.8416~~\underline{0.9009}	 & 	
\underline{0.9614}~~0.9241~~0.9081~~0.8571	 & 	0.9560~~0.9554~~0.9481~~0.9475	 & 	
\textbf{0.9962}~~0.9902~~0.9882~~0.9692	 \\

\midrule[0.5pt]

\multirow{4}{*}{\rotatebox{90}{\fontsize{5.0pt}{\baselineskip}\selectfont TS-Conv-LSTM}}	 
&    MAE     &  	
0.0368~~0.0396~~0.0461~~0.0372	 & 	
0.0363~~0.0532~~0.0501~~0.0647	 &
0.0313~~\underline{0.0294}~~0.0314~~0.0308	 &
\underline{0.0084}~~\underline{0.0109}~~\underline{0.0127}~~\underline{0.0262}	 \\

&    RMSE    &  	
0.0476~~0.0537~~0.0623~~0.0486	 & 	
0.0784~~0.0978~~0.1104~~0.1331	 &
0.0484~~0.0463~~\underline{0.0473}~~0.0482	 &
\underline{0.0110}~~\underline{0.0139}~~0.0165~~\underline{0.0338}	\\
  
&    RSE     &  	
0.5067~~0.5721~~0.6633~~0.5179	 &
0.3153~~0.3934~~0.4442~~0.5357	 & 	
0.4455~~0.4263~~\underline{0.4351}~~0.4439	 & 	
\underline{0.1103}~~\underline{0.1403}~~0.1658~~\underline{0.3406}	 \\
  
&    CORR     &  	
0.8696~~0.8533~~0.8146~~0.8561	 & 	
0.9545~~0.9318~~0.9082~~0.8653	 & 	
0.9548~~0.9546~~\underline{0.9544}~~\underline{0.9573}	 & 	
\textbf{0.9962}~~\underline{0.9932}~~\textbf{0.9891}~~\textbf{0.9790}  \\

\midrule[0.5pt]

\multirow{4}{*}{\rotatebox{90}{TCLN}}	
&    MAE     & 	
0.0361~~0.0399~~0.0402~~0.0373	 & 
\textbf{0.0320}~~0.0575~~0.0574~~0.0654	 &
0.0287~~0.0302~~0.0395~~0.0329	 
& 	0.0141~~0.0112~~0.0240~~0.0355	 \\

&    RMSE     &  	
0.0486~~0.0508~~0.0524~~\underline{0.0485}	 & 	
\underline{0.0711}~~0.0974~~0.1156~~0.1339	 &
0.0448~~\textbf{0.0441}~~0.0560~~0.0493	  &
0.0174~~0.0147~~0.0290~~0.0414	 \\
  
&    RSE     &  	
0.5176~~0.5405~~0.5584~~\underline{0.5162}	 & 	
\underline{0.2859}~~0.3918~~0.4653~~05388	 &
0.4126~~\textbf{0.4062}~~0.5156~~0.4540	 & 	0.1751~~0.1482~~0.2916~~0.4168	  \\
  
&    CORR     &  	
0.8889~~0.8489~~0.8540~~0.8645	 & 	
0.9596~~\underline{0.9330}~~0.9017~~0.8672	 & 
0.9540~~0.9504~~\textbf{0.9589}~~0.9563	 &
0.9944~~0.9907~~0.9840~~0.9622	 \\

\midrule[0.5pt]

\multirow{4}{*}{\rotatebox{90}{Ours}}	 
&    MAE     &  	
\textbf{0.0253}~~\textbf{0.0288}~~\textbf{0.0294}~~\textbf{0.0293}	 &
0.0332~~\underline{0.0447}~~\textbf{0.0495}~~\underline{0.0595}	 &
\textbf{0.0257}~~\textbf{0.0285}~~\textbf{0.0297}~~\textbf{0.0291}	 &
\textbf{0.0070}~~\textbf{0.0098}~~\textbf{0.0122}~~\textbf{0.0218}	 \\

&    RMSE     &  	
\textbf{0.0333}~~\textbf{0.0385}~~\textbf{0.0396}~~\textbf{0.0401}	 & 
\textbf{0.0681}~~\textbf{0.0916}~~\textbf{0.1030}~~\textbf{0.1226}   &
\underline{0.0406}~~\underline{0.0448}~~\textbf{0.0438}~~\textbf{0.0435}   &	
\textbf{0.0098}~~\textbf{0.0131}~~\textbf{0.0154}~~\textbf{0.0283}	 \\
  
&    RSE     &  	
\textbf{0.3567}~~\textbf{0.4122}~~\textbf{0.4239}~~\textbf{0.4292}	 &
\textbf{0.2750}~~\textbf{0.3696}~~\textbf{0.4157}~~\textbf{0.4947}   &
\underline{0.3775}~~\underline{0.4165}~~\textbf{0.4066}~~\textbf{0.4041}   &
\textbf{0.1009}~~\textbf{0.1345}~~\textbf{0.1585}~~\textbf{0.2912}	 \\
  
&    CORR     &  	
\textbf{0.9374}~~\textbf{0.9240}~~\textbf{0.9063}~~\textbf{0.9080}	 &
\textbf{0.9620}~~\textbf{0.9336}~~\textbf{0.9146}~~\textbf{0.8838}   &
\underline{0.9608}~~\textbf{0.9568}~~0.9499~~\textbf{0.9604}   &	
0.9957~~\textbf{0.9933}~~\underline{0.9889}~~\underline{0.9734}	   \\

\bottomrule[1.0pt]

\end{tabular}
}
\end{table*}

In this experiment, we set the same hyperparameters for different public datasets’ prediction tasks to show that the DA-SPS model has good generalization ability. 
The threshold of Spearman Correlation Coefficient, $\rho^{\prime}$, is set as $0.5$. 
The window size $t$ is set as $96$. 
The length of each subsequence after patching and the size of the hidden layer units in LSTM, P-Conv-LSTM, and L-Attention model are set as $24$ and $64$, respectively. 
The training process involves $200$ epochs and each iteration $batch$ is $64$. 
The initial learning rate $lr$ is $0.001$ and the learning rate decay ${lr}_{g}$ is $0.9$. For a detailed description of the training process, please refer to Algorithm~\ref{algorithm:Training Process_THZhang}:

\begin{algorithm}[htbp]
    \normalsize  
    \caption{Training Process}
    \label{algorithm:Training Process_THZhang}
    \textbf{Require}: $epoch$; $lr$; ${lr}_{g}$, the learning rate decay, default to 0.9; $AT$, the target variable sequence; $ET$, the extraneous variable sequences; $T_{P}$, the prediction value; $label$, the ground truth

    \begin{algorithmic}[1]
        \State $NormET \leftarrow (ET-ET_{min})/({ET_{max}-ET_{min}})$
        \State $NormAT \leftarrow (AT-AT_{min})/(AT_{max}-AT_{min})$
        \State $AT_{se}$, $AT_{tr}$ $\leftarrow$ \textbf{Model.SSA}($NormAT$)
        \State $ET_{cor}$ $\leftarrow$ \textbf{Model.Spearman}($NormET$)
        
        \For{$i=1$ \textbf{to} $epoch$}:
            \State \textbf{if} $i$\%20 == 0 : $lr \leftarrow lr \times {lr}_{g}$
        \EndFor
        
        \State $T_{P}$ $\leftarrow$ \textbf{Model.Forward}($ET_{cor}$, $AT_{se}$, $AT_{te}$)
        \State $loss$ = \textbf{MAE}($T_{P}$, $label$)
        \State \textbf{Model.Backward}($loss$, $lr$)

    \end{algorithmic}
            
\end{algorithm}

To ensure comparability of the results, the baselines' hyperparameters are set uniformly as follows: $epoch=200$, $batch=64$, $lr=0.001$ and ${lr}_g=0.9$.
We validated the above eight models under the different prediction tasks, $\operatorname{horizon}=\{3, 6, 12, 24\}$.
The performance of each model at different prediction tasks is recorded in TABLE~\ref{tab:exp.mainResults}.
It should be noted that the performance evaluation of these models is all based on the normalized data.

\begin{figure*}[htbp]

    \subfigure[Electricity MAE Curves] 
	{
		\begin{minipage}[t]{8cm}
			\centering          
			\includegraphics[scale=0.22]{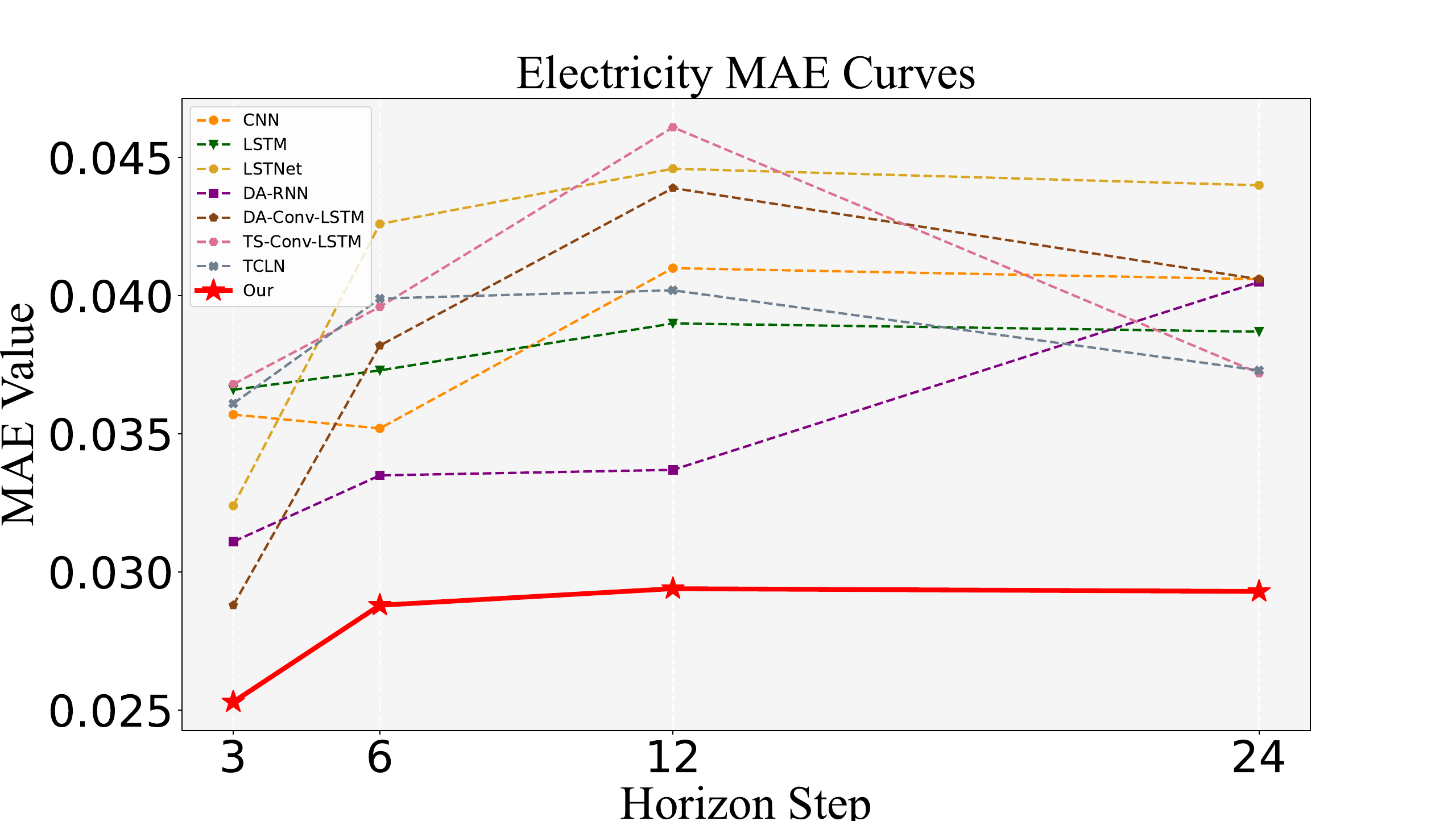}  
		\end{minipage}
	}
\hspace{2mm}
     \subfigure[Solar MAE Curves] 
    	{
    		\begin{minipage}[t]{8cm}
    			\centering          
    			\includegraphics[scale=0.22]{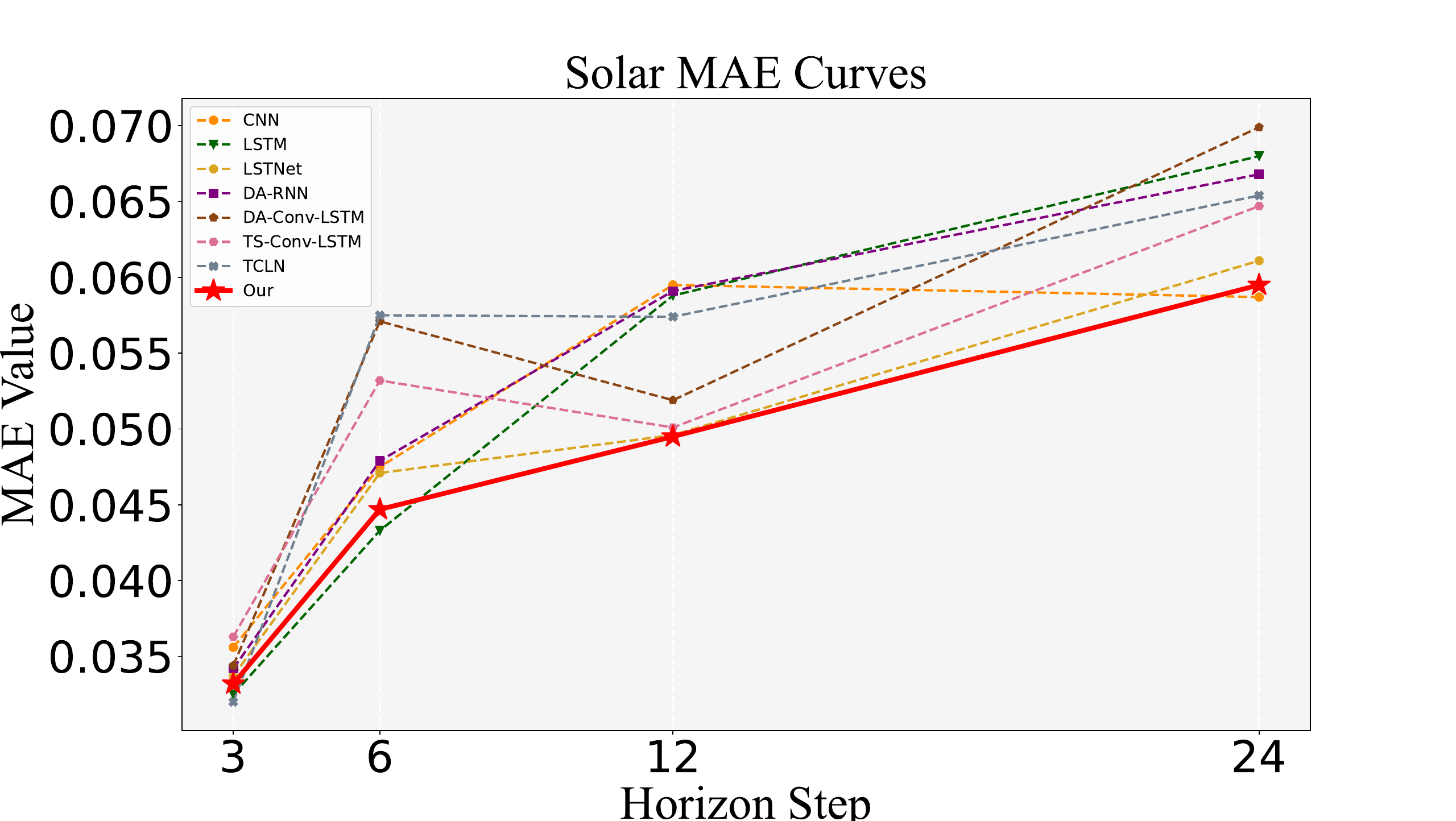}  
    		\end{minipage}
    	}

    \subfigure[Traffic MAE Curves] 
	{
		\begin{minipage}[t]{8cm}
			\centering          
			\includegraphics[scale=0.22]{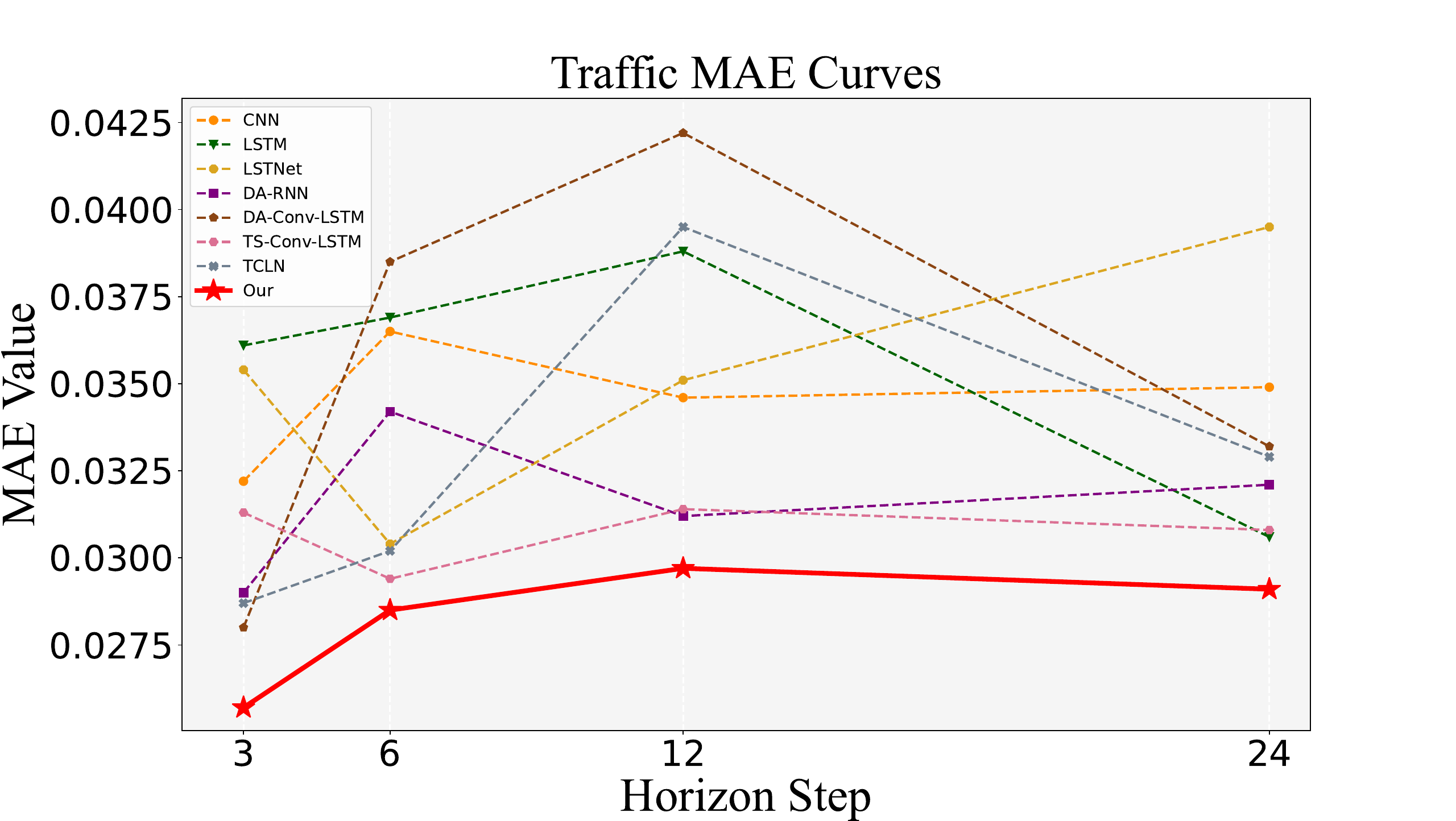}  
		\end{minipage}
	}
\hspace{2mm}
     \subfigure[Exchange MAE Curves] 
    	{
    		\begin{minipage}[t]{8cm}
    			\centering          
    			\includegraphics[scale=0.22]{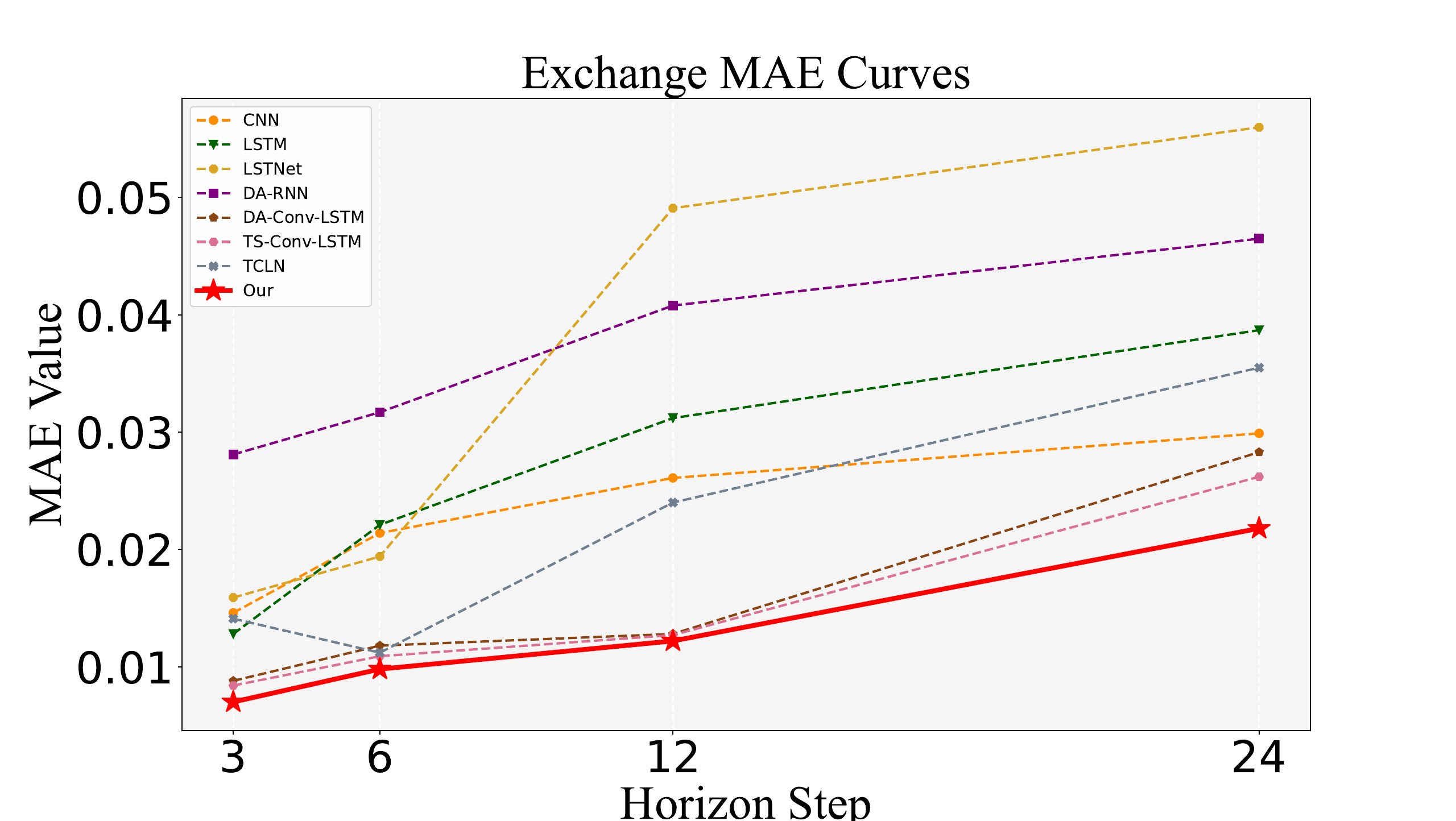}  
    		\end{minipage}
    	}
    \caption{MAE curves for each model across Electricity, Solar, Traffic, and Exchange datasets for various prediction tasks.}
    \label{fig:model_MAE}
\end{figure*}

According to TABLE~\ref{tab:exp.mainResults}, our model shows superior performance in various tasks.
Fig.~\ref{fig:model_MAE} presents the MAE curves of each model under diverse datasets, and it can be observed that the MAE values of the DA-SPS are lower than those of the other baselines.
It suggests that the DA-SPS has a smaller estimation error and more stable performance when making forecasts. 

\begin{figure*}[htbp]

    \subfigure[Electricity] 
	{
		\begin{minipage}[t]{8cm}
			\centering          
			\includegraphics[scale=0.22]{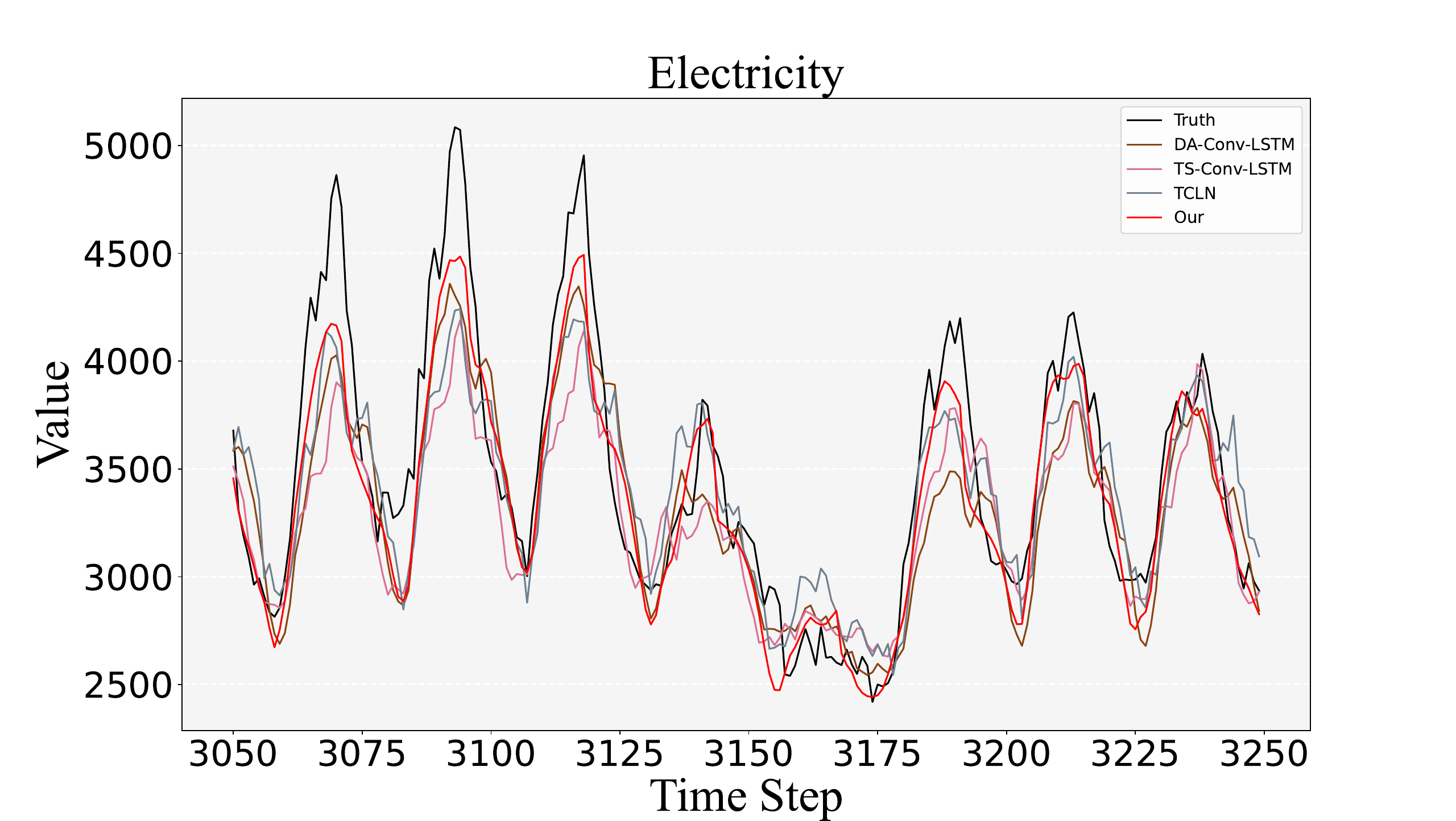}  
		\end{minipage}
	}
\hspace{2mm}
     \subfigure[Solar] 
    	{
    		\begin{minipage}[t]{8cm}
    			\centering          
    			\includegraphics[scale=0.22]{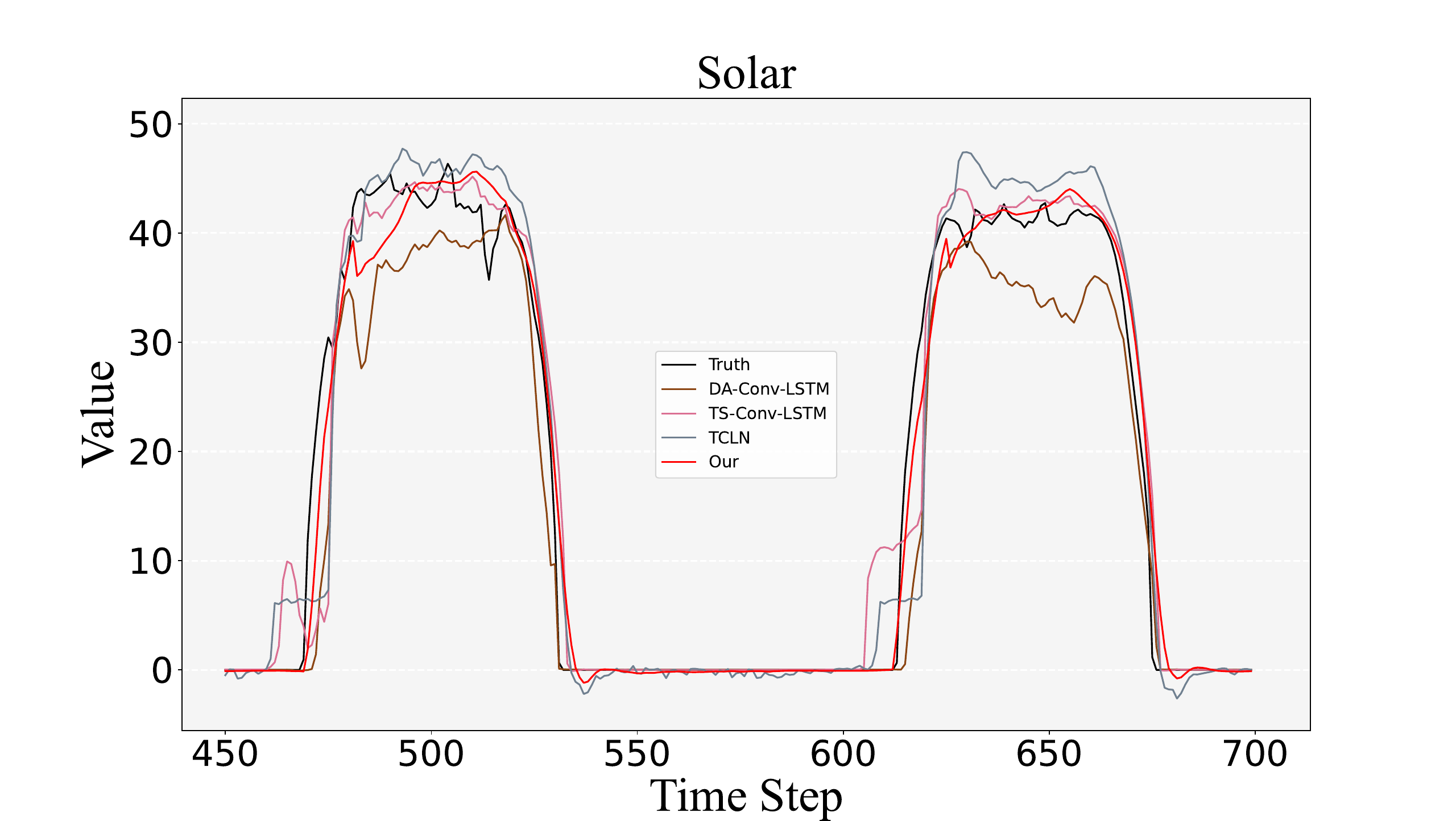}  
    		\end{minipage}
    	}

    \subfigure[Traffic] 
	{
		\begin{minipage}[t]{8cm}
			\centering          
			\includegraphics[scale=0.22]{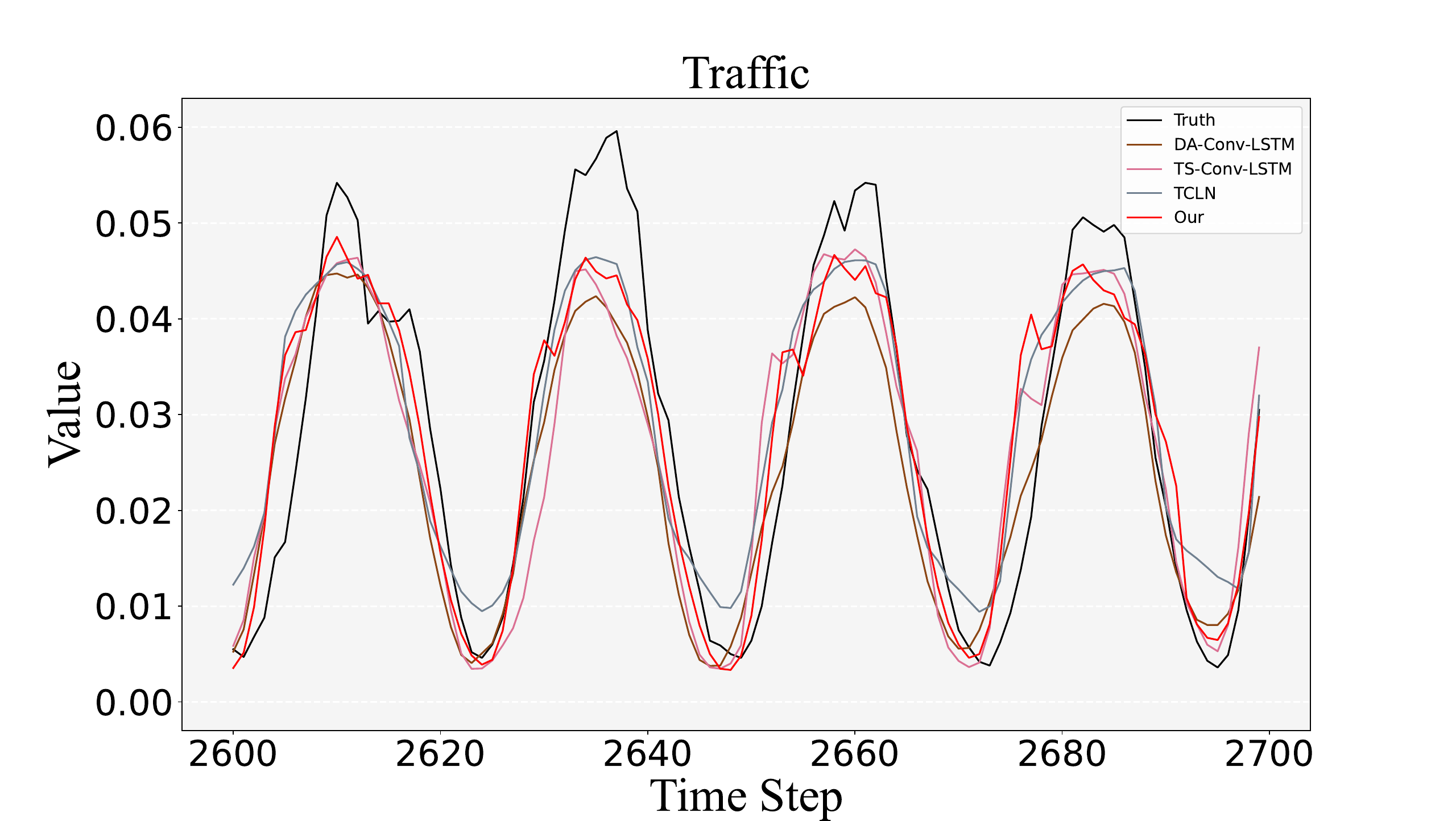}  
		\end{minipage}
	}
\hspace{2mm}
     \subfigure[Exchange] 
    	{
    		\begin{minipage}[t]{8cm}
    			\centering          
    			\includegraphics[scale=0.22]{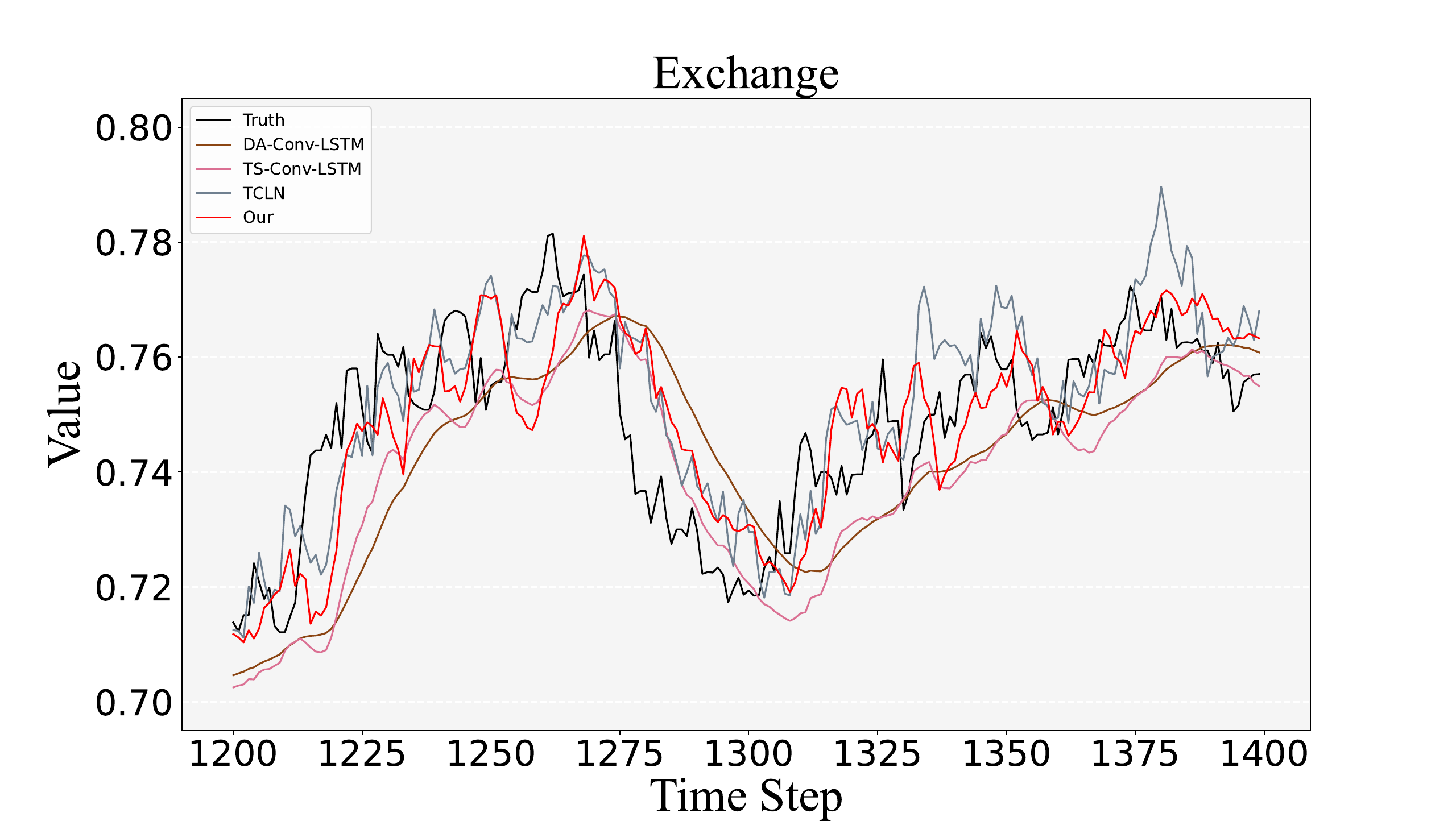}  
    		\end{minipage}
    	}
    \caption{Prediction curves for different models in the $\operatorname{horizon}=6$ task.}
    \label{fig:predict_pubilcdata}
\end{figure*}

To further intuitively compare the predictive performance of various models, we present the prediction curves of each model for the task, $\operatorname{horizon}=6$, on four datasets, as shown in Fig.~\ref{fig:predict_pubilcdata}.
From the figure, it can be observed that the prediction curve of the DA-SPS is more in line with true values than the other baselines. 
It significantly indicates the superiority and reliability of the DA-SPS in MTSF tasks.

\subsection{Laptop Board Yield Prediction}

In addition to validating the DA-SPS on a series of public datasets, we further apply it to the LBY dataset to accurately predict the yield variation of different function test items on the Laptop PCB board.
Yield prediction can help production lines develop appropriate testing strategies. When the yield of a particular function test item shows a downward trend, the production lines can adjust the testing frequency of this test item in advance, thus effectively preventing the unqualified PCB boards from entering the market.

\begin{table}[htbp]
\centering
\caption{Names and Average Testing Time of Function Test Items.}
\label{tab:privatedata_descrip_THZhang}
\fontsize{9pt}{9pt}\selectfont
\centering
\begin{tabular}{c|c}
\toprule [2.0pt]
Function Test Item Names &  Average Testing Time  \\
\midrule [1.0pt]
SmartKeyboard & 7.797s \\
DOCKMAC & 1.188s \\
Brightness & 2.391s \\
CPUCheck & 0.312s \\
Sensor & 8.422s \\
CamerID & 1.453s \\
RAMCHK & 0.187s\\
\bottomrule [2.0pt]  
\end{tabular}
\end{table}

In this experiment, we use a private dataset provided by a well-known laptop manufacturer, which contained 9,577 records on the yields of seven different test items in the function test process of Laptop PCB boards.
The names of each of these test items and the average testing time are shown in TABLE~\ref{tab:privatedata_descrip_THZhang}.

Considering the complexity of the PCB boards' manufacturing process, a single component failure can possibly cause multiple functions of the motherboard to malfunction. Furthermore, the failure of the component often indicates potential problems such as faults or ageing in the production equipment, which can typically affect the quality of an entire batch of components.
Therefore, in predicting the yield of a particular test item, we must not only extract and analyze the trend and seasonal variations of that sequence, but also take full advantage of the correlations between it and the other test items. By exploiting the yield changes of related test item sequences, we can improve the accuracy of the predictions.

\begin{figure}[htbp]
	\centering
	\includegraphics[width=\linewidth]{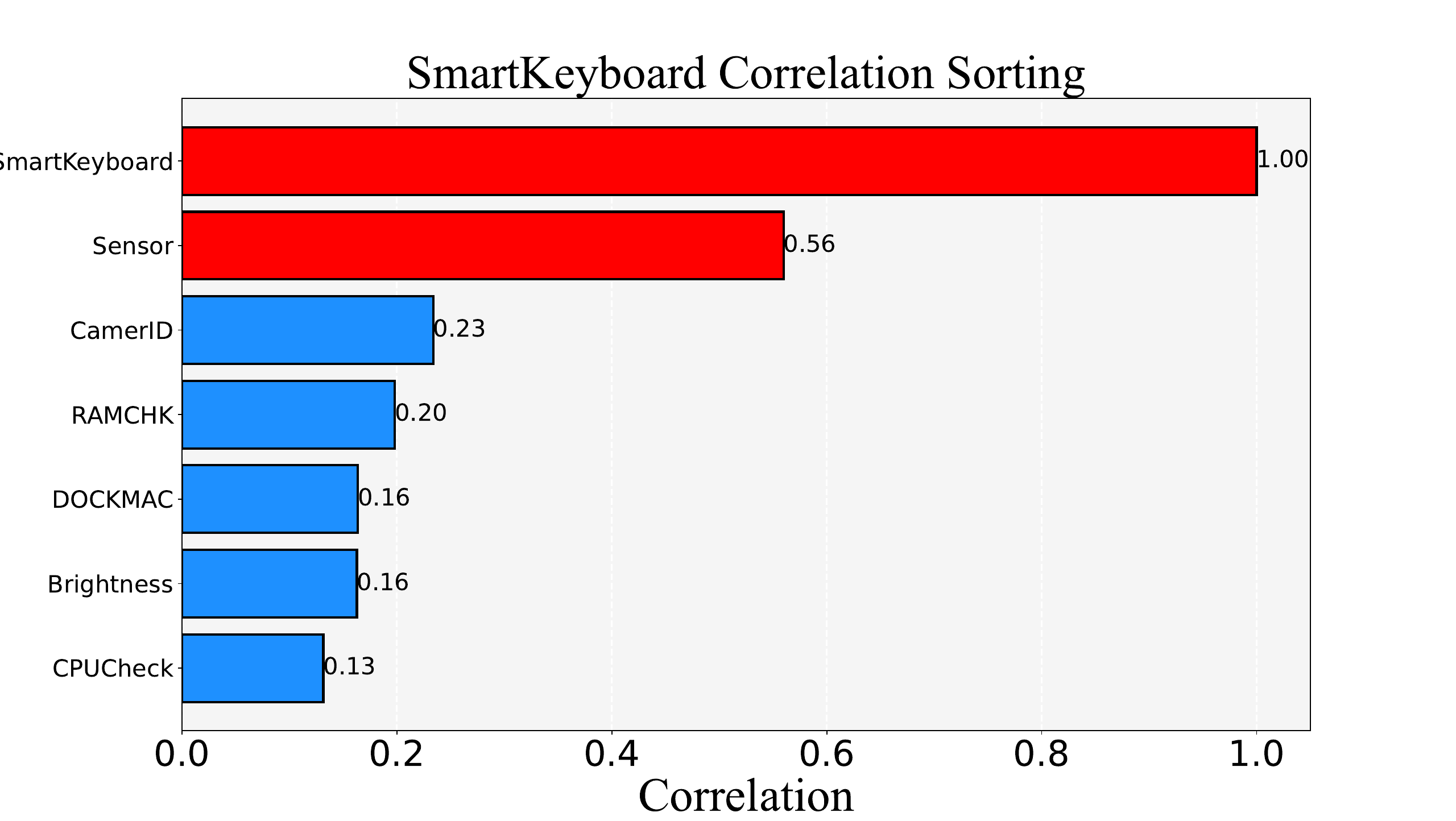}
	\caption{
The Spearman Correlation Coefficients between SmartKeyboard and other test items, ranked by absolute value from the strongest to the weakest.}
	\label{fig:SmartKeyboard Correlation & Sorting}
\end{figure}

We choose SmartKeyboard as the target item for prediction and utilize the DA-SPS model's Spearman Correlation Analysis module to calculate its correlation coefficient with other test items. 
We set the threshold at $\rho^{\prime}=0.5$ and the results are shown in Fig.~\ref{fig:SmartKeyboard Correlation & Sorting}.

Fig.~\ref{fig:SmartKeyboard Correlation & Sorting} clearly shows that there is a strong correlation between the SmartKeyboard and the Sensor. Therefore, the DA-SPS selectively incorporates the Sensor as a key auxiliary variable for enhancing the yield prediction accuracy of the target, thus effectively mitigating the potential interference from weakly correlated items. 

\begin{figure}[htbp]
	\centering
	\includegraphics[width=\linewidth]{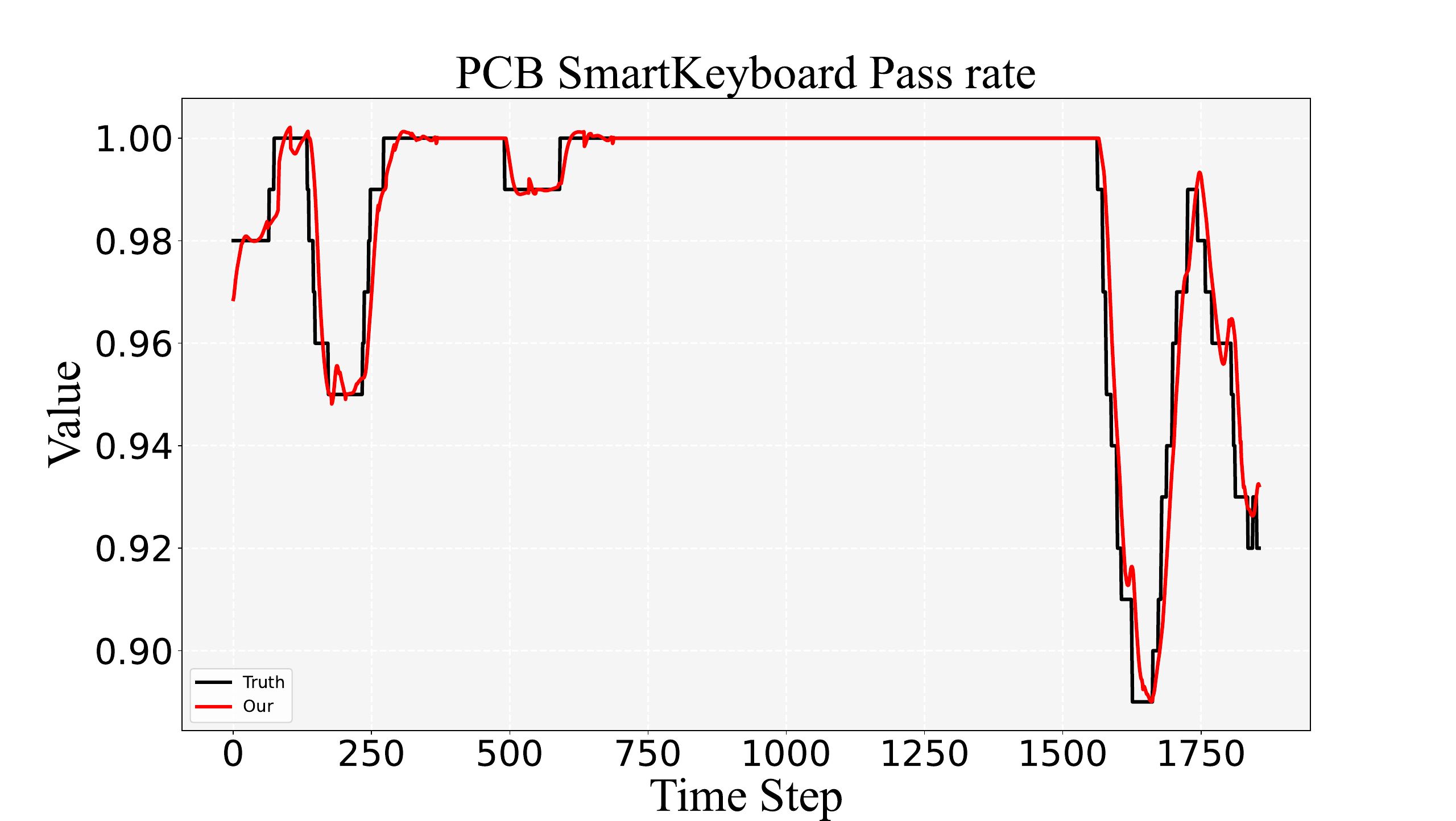}
	\caption{Prediction result curve for the yield of SmartKeyboard Item at $\operatorname{horizon}=3$.}
	\label{fig:SmartKeyboard_predict_THZhang}
\end{figure}

TABLE~\ref{tab:PCB_Function_Test_THZhang} shows the prediction results for the target item across the four tasks, $\operatorname{horizon}=\{3, 6, 12, 24\}$. Taking the horizon=3 task as an instance, the prediction curve is shown in Fig.~\ref{fig:SmartKeyboard_predict_THZhang}. Despite deviations between the predicted and true yields, the general performance of the model is satisfactory, indicating a competent predictive capability.

\begin{table}[htbp]
\centering
\caption{Prediction Results of SmartKeyboard Item for Various Horizon Tasks.}
\label{tab:PCB_Function_Test_THZhang}
\fontsize{8pt}{8pt}\selectfont
\centering
\begin{tabular}{c|c|c|c|c|c}
\toprule[1.0pt]
\multicolumn{2}{c|}{Metric}         & {MAE}  & {RMSE} & {RSE} & {CORR}  \\
\midrule[1.0pt]
\multirow{4}{*}{\rotatebox{90}{horizon}}	 
& 	3	 &  	0.0025 & 0.0059 & 0.2305 & 0.9733 \\
& 	6	 &  	0.0042 & 0.0087 & 0.3412 & 0.9466 \\
& 	12	 &  	0.0042 & 0.0091 & 0.3586 & 0.9404 \\
& 	24	 &  	0.0058 & 0.0121 & 0.4755 & 0.9041 \\
\bottomrule[1.0pt]
\end{tabular}
\end{table}

\begin{table*}[htbp]
\caption{Ablation study results of DA-SPS on Electricity, Solar, Traffic, Exchange, and LBY under $\operatorname{horizon=12}$.}
\label{tab:Ablation test in DA-SPS}
	\vspace{3pt}
	\renewcommand{\arraystretch}{2} 
	\centering
	\resizebox{\textwidth}{!}{

			\renewcommand{\multirowsetup}{\centering}
			\setlength{\tabcolsep}{2pt}
			\begin{tabular}{c|c|cccc|cccc|cccc|cccc|cccc}
				\toprule
				\multicolumn{2}{c}{\multirow{1}{*}{Datasets}}  &
				\multicolumn{4}{c}{\rotatebox{0}{\scalebox{0.8}{Electricity}}} &
				\multicolumn{4}{c}{\rotatebox{0}{\scalebox{0.8}{Solar}}} &
				\multicolumn{4}{c}{\rotatebox{0}{\scalebox{0.8}{{Traffic}}}} &
				\multicolumn{4}{c}{\rotatebox{0}{\scalebox{0.8}{Exchange}}}  &
				\multicolumn{4}{c}{\rotatebox{0}{\scalebox{0.8}{LBY}}}   \\
				\cmidrule(lr){1-2}
				\cmidrule(lr){3-6} \cmidrule(lr){7-10}\cmidrule(lr){11-14} \cmidrule(lr){15-18} \cmidrule(lr){19-22}
				\multicolumn{2}{c|}{Metrics}  & \scalebox{0.78}{MAE} & \scalebox{0.78}{RMSE}  & \scalebox{0.78}{RSE} & \scalebox{0.78}{CORR}  & \scalebox{0.78}{MAE} & \scalebox{0.78}{RMSE}  & \scalebox{0.78}{RSE} & \scalebox{0.78}{CORR} & \scalebox{0.78}{MAE} & \scalebox{0.78}{RMSE}  & \scalebox{0.78}{RSE} & \scalebox{0.78}{CORR} & \scalebox{0.78}{MAE} & \scalebox{0.78}{RMSE}  & \scalebox{0.78}{RSE} & \scalebox{0.78}{CORR} & \scalebox{0.78}{MAE} & \scalebox{0.78}{RMSE}  & \scalebox{0.78}{RSE} & \scalebox{0.78}{CORR}  \\
				\toprule

    \multirow{4}{*}{\rotatebox{90}{\scalebox{0.95}{Ablation}}}
				& \scalebox{0.78}{DA-SPS} & {\scalebox{0.78}{\textbf{0.0294}}} &{\scalebox{0.78}{\textbf{0.0396}}} & \scalebox{0.78}{\textbf{0.4239}} &\scalebox{0.78}{\textbf{0.9063}} & {\scalebox{0.78}{\textbf{0.0495}}} &{\scalebox{0.78}{\textbf{0.1030}}} & \scalebox{0.78}{\textbf{0.4157}} &\scalebox{0.78}{\textbf{0.9146}} & {\scalebox{0.78}{\textbf{0.0297}}} &{\scalebox{0.78}{\textbf{0.0438}}} & \scalebox{0.78}{\textbf{0.4066}} &\scalebox{0.78}{\textbf{0.9499}} & {\scalebox{0.78}{\textbf{0.0122}}} &{\scalebox{0.78}{\textbf{0.0154}}} & \scalebox{0.78}{\textbf{0.1585}} &\scalebox{0.78}{\textbf{0.9889}} & {\scalebox{0.78}{\textbf{0.0042}}} &{\scalebox{0.78}{\textbf{0.0091}}} & \scalebox{0.78}
                {\textbf{0.3586}} &
                \scalebox{0.78}
                {\textbf{0.9404}} \\
    
				& \scalebox{0.78}{W/O SSA} & {\scalebox{0.78}{0.0309}} &{\scalebox{0.78}{0.0420}} & \scalebox{0.78}{0.4497} &\scalebox{0.78}{0.8950} & {\scalebox{0.78}{0.0592}} &{\scalebox{0.78}{0.1165}} & \scalebox{0.78}{0.4703} &\scalebox{0.78}{0.8982} & {\scalebox{0.78}{0.0340}} &{\scalebox{0.78}{0.0468}} & \scalebox{0.78}{0.4346} &\scalebox{0.78}{0.9347} & {\scalebox{0.78}{0.0258}} &{\scalebox{0.78}{0.0309}} & \scalebox{0.78}{0.3173} &\scalebox{0.78}{0.9771} & {\scalebox{0.78}{0.0054}} &{\scalebox{0.78}{0.0110}} & \scalebox{0.78}{0.4342} &\scalebox{0.78}{0.9036}    \\
    
				& \scalebox{0.78}{W/O P-Conv-LSTM}& {\scalebox{0.78}{0.0337}} &{\scalebox{0.78}{0.0472}} & \scalebox{0.78}{0.5058} &\scalebox{0.78}{0.8797} & {\scalebox{0.78}{0.0582}} &{\scalebox{0.78}{0.1155}} & \scalebox{0.78}{0.4661} &\scalebox{0.78}{0.8905} & {\scalebox{0.78}{0.0325}} &{\scalebox{0.78}{0.0446}} & \scalebox{0.78}{0.4148} &\scalebox{0.78}{0.9341} & {\scalebox{0.78}{0.0352}} &{\scalebox{0.78}{0.0502}} & \scalebox{0.78}{0.5159} &\scalebox{0.78}{0.9551} & {\scalebox{0.78}{0.0090}} &{\scalebox{0.78}{0.0136}} & \scalebox{0.78}{0.5358} &\scalebox{0.78}{0.8818}  \\
    
				& \scalebox{0.78}{W/O Spearman} & {\scalebox{0.78}{0.0338}} &{\scalebox{0.78}{0.0471}} & \scalebox{0.78}{0.5046} &\scalebox{0.78}{0.8853} & {\scalebox{0.78}{0.0533}} &{\scalebox{0.78}{0.1125}} & \scalebox{0.78}{0.4540} &\scalebox{0.78}{0.9023} & {\scalebox{0.78}{0.0366}} &{\scalebox{0.78}{0.0536}} & \scalebox{0.78}{0.4977} &\scalebox{0.78}{0.9456} & {\scalebox{0.78}{0.0302}} &{\scalebox{0.78}{0.0357}} & \scalebox{0.78}{0.3671} &\scalebox{0.78}{0.9731} & {\scalebox{0.78}{0.0057}} &{\scalebox{0.78}{0.0119}} & \scalebox{0.78}{0.4674} &\scalebox{0.78}{0.8998}  \\

				\bottomrule
			\end{tabular}

	}
\end{table*}

\begin{table*}[htbp]
\caption{Comparative Analysis of DA-SPS Model with SSA and STL Module at $\operatorname{Horizon} = 12$.}
\label{tab:different in SVD and STL THZhang}
	\vspace{3pt}
	\renewcommand{\arraystretch}{2} 
	\centering
	\resizebox{\textwidth}{!}{
			\begin{huge}
			\renewcommand{\multirowsetup}{\centering}
			\setlength{\tabcolsep}{4pt}
			\begin{tabular}{c|c|cccc|cccc|cccc|cccc|cccc}
				\toprule
				\multicolumn{2}{c}{\multirow{1}{*}{Datasets}}  &
				\multicolumn{4}{c}{\rotatebox{0}{\scalebox{0.8}{Electricity}}} &
				\multicolumn{4}{c}{\rotatebox{0}{\scalebox{0.8}{Solar}}} &
				\multicolumn{4}{c}{\rotatebox{0}{\scalebox{0.8}{{Traffic}}}} &
				\multicolumn{4}{c}{\rotatebox{0}{\scalebox{0.8}{Exchange}}}  &
				\multicolumn{4}{c}{\rotatebox{0}{\scalebox{0.8}{LBY}}}   \\
				\cmidrule(lr){1-2}
				\cmidrule(lr){3-6} \cmidrule(lr){7-10}\cmidrule(lr){11-14} \cmidrule(lr){15-18} \cmidrule(lr){19-22}
				\multicolumn{2}{c|}{Metrics}  & \scalebox{0.78}{MAE} & \scalebox{0.78}{RMSE}  & \scalebox{0.78}{RSE} & \scalebox{0.78}{CORR}  & \scalebox{0.78}{MAE} & \scalebox{0.78}{RMSE}  & \scalebox{0.78}{RSE} & \scalebox{0.78}{CORR} & \scalebox{0.78}{MAE} & \scalebox{0.78}{RMSE}  & \scalebox{0.78}{RSE} & \scalebox{0.78}{CORR} & \scalebox{0.78}{MAE} & \scalebox{0.78}{RMSE}  & \scalebox{0.78}{RSE} & \scalebox{0.78}{CORR} & \scalebox{0.78}{MAE} & \scalebox{0.78}{RMSE}  & \scalebox{0.78}{RSE} & \scalebox{0.78}{CORR}  \\
				\toprule

    \multirow{2}{*}{\rotatebox{90}{\scalebox{0.95}{Contrast}}}
				& \scalebox{0.78}{SSA} & {\scalebox{0.78}{\textbf{0.0294}}} &{\scalebox{0.78}{\textbf{0.0396}}} & \scalebox{0.78}{\textbf{0.4239}} &\scalebox{0.78}{\textbf{0.9063}} & {\scalebox{0.78}{\textbf{0.0495}}} &{\scalebox{0.78}{\textbf{0.1030}}} & \scalebox{0.78}{\textbf{0.4157}} &\scalebox{0.78}{\textbf{0.9146}} & {\scalebox{0.78}{\textbf{0.0297}}} &{\scalebox{0.78}{\textbf{0.0438}}} & \scalebox{0.78}{\textbf{0.4066}} &\scalebox{0.78}{\textbf{0.9499}} & {\scalebox{0.78}{\textbf{0.0122}}} &{\scalebox{0.78}{\textbf{0.0154}}} & \scalebox{0.78}{\textbf{0.1585}} &\scalebox{0.78}{\textbf{0.9889}} & {\scalebox{0.78}{\textbf{0.0042}}} &{\scalebox{0.78}{\textbf{0.0091}}} & \scalebox{0.78}
                {\textbf{0.3586}} &
                \scalebox{0.78}
                {\textbf{0.9404}} \\
    
				& \scalebox{0.78}{STL} & {\scalebox{0.78}{0.0338}} &{\scalebox{0.78}{0.0458}} & \scalebox{0.78}{0.4905} &\scalebox{0.78}{0.8966} & {\scalebox{0.78}{0.0618}} &{\scalebox{0.78}{0.1189}} & \scalebox{0.78}{0.4798} &\scalebox{0.78}{0.8885} & {\scalebox{0.78}{0.0313}} &{\scalebox{0.78}{0.0452}} & \scalebox{0.78}{0.4202} &\scalebox{0.78}{0.9363} & {\scalebox{0.78}{0.0172}} &{\scalebox{0.78}{0.0216}} & \scalebox{0.78}{0.2224} &\scalebox{0.78}{0.9810} & {\scalebox{0.78}{0.0065}} &{\scalebox{0.78}{0.0096}} & \scalebox{0.78}{0.3790} &\scalebox{0.78}{0.9296}  \\

				\bottomrule
			\end{tabular}
		\end{huge}
	}
\end{table*}

\subsection{Ablation Study}

In this part, we demonstrate the validity of each module of the DA-SPS by designing ablation studies. 
We prepare the following ablation studies according to the function of each module:

\begin{itemize}
    \item[$\bullet$] \textbf{W/O SSA}: The DA-SPS model removes the SSA module, which is mainly responsible for decomposing the target variable sequence.
    \item[$\bullet$] \textbf{W/O P-Conv-LSTM}: The LSTM is used instead of the P-Conv-LSTM module that is responsible for the feature extraction from the trend sequence of the target variable.
    \item[$\bullet$] \textbf{W/O Spearman}: The DA-SPS removes the Spearman correlation analysis module, which primarily filters the extraneous variables' sequences.
\end{itemize}

We choose the Electricity, Solar, Traffic, Exchange and LBY datasets for validation under $\operatorname{horizon}=12$. To ensure the reliability of the ablation studies' results, the hyperparameters of each ablation study are consistent with those of the DA-SPS model under the same task. The ablation studies still choose MAE, RMSE, RSE and CORR as metrics, and the results are shown in TABLE~\ref{tab:Ablation test in DA-SPS}, which have the following notable points:

\begin{itemize}
    \item[$\bullet$] The results of three ablation studies (W/O SSA, W/O P-Conv-LSTM, W/O Spearman) all show some degree of degradation compared to the original model, which is especially obvious in the Electricity, Solar, and LBY datasets.
    \item[$\bullet$] Replacing P-Conv-LSTM with LSTM (W/O P-Conv-LSTM) had the most significant effect on the model performance, indicating the crucial role of the P-Conv-LSTM module.
    \item[$\bullet$] The significance of the Spearman module varies across different datasets, with the most significant impact in the LBY dataset.
    
\end{itemize}

As for why the P-Conv-LSTM module plays the most important role, we explain it as follows: Considering that the seasonality component exhibits a repetitive and stable pattern, extracting and processing information from the trend component is crucial for accurate future prediction of the target. The P-Conv-LSTM module is just adept at analysing the local data of the trend component, offering a more comprehensive understanding of the underlying patterns. Therefore, if this module is replaced, it will directly impact the effectiveness of the DA-SPS.

For the SSA and Spearman correlation analysis modules, the results of each dataset also decrease after removing. The effect is most significant on the LBY dataset, indicating that in the yield prediction task of the Laptop PCB board, decomposing the target variable sequence as well as filtering the extraneous variables' sequences can effectively improve the prediction accuracy.

To highlight the benefits of the SSA method, we also conduct a comparative analysis, where the SSA module in DA-SPS is substituted with STL (Seasonal and Trend decomposition using Loess)\cite{cleveland1990stl} module. STL module decomposes sequences into seasonality and trend components through a series of loess smoothers. 
STL has been proven effective and is widely used in models such as PCDformer\cite{ma2024multivariate} and Autoformer\cite{Wu2021}.

The design of the STL module is based on the structure of the Series Decomposition Block in Autoformer, which is formulated as shown below:
\begin{equation}\label{eq: STL_eq_THZhang}
    \begin{aligned}
    &X^{gt} = \operatorname{AvgPool}(\operatorname{Padding}(X))\\
    &X^{gs} = X-X^{gt},\\
    \end{aligned}
\end{equation}
where $\operatorname{AvgPool(\cdot)}$ denotes the moving average algorithm with the padding operation to keep the length of sequences unchanged.
Subsequently, the impact of the STL and SSA modules on the DA-SPS is evaluated under the $horizon = 12$, with the performance presented in TABLE~\ref{tab:different in SVD and STL THZhang}.

The results indicate that the SSA module outperforms the STL module on all datasets. This is primarily due to the SSA module's ability to mitigate the impact of noise on sequence analysis by retaining only the trend and seasonality sequences after sequence decomposition, while the STL module does not remove the noise. 

\section{Conclusion}  \label{conclusion}

In this paper, we propose a new prediction model, DA-SPS, for MTSF tasks.
The model mainly comprises two stages, TVPS and EVPS, to process the target variable and the extraneous variables, separately. 
TVPS consists of SSA, LSTM, and P-Conv-LSTM modules, which strongly enhance the extraction of temporal features from the target variable. EVPS contains Spearman correlation analysis and L-Attention modules to minimize the impact of irrelevant variables while making more comprehensive use of the information contained in relevant variables' sequences.
Experimental results on both public and private datasets show that the DA-SPS outperforms the state-of-the-art models and has greater potential in industrial datasets with complex situations.

\bibliographystyle{IEEEtran}
\bibliography{IEEEabrv,reference}

\end{document}